# Property-Guided Inverse Design of Metal-Organic Frameworks Using Quantum Natural Language Processing

*Shinyoung Kang, Jihan Kim**

Department of Chemical and Biomolecular Engineering, Korea Advanced Institute of Science and Technology

291 Daehak-ro, Yuseong-gu, Daejeon 34141, Republic of Korea

*E-mail: jihankim@kaist.ac.kr, Phone: +82-42-350-7311 (Corresponding author)

# Abstract

In this study, we explore the potential of using quantum natural language processing (QNLP) for property-guided inverse design of metal-organic frameworks (MOFs) with targeted properties. Specifically, by analyzing 450 hypothetical MOF structures consisting of 3 topologies, 10 metal nodes and 15 organic ligands, we categorize these structures into four distinct classes for pore volume and $CO_2$ Henry's constant values. We then compare various QNLP models (i.e. the bag-of-words, DisCoCat (Distributional Compositional Categorical), and sequence-based models) to identify the most effective approach to process the MOF dataset. Using a classical simulator provided by the IBM Qiskit, the bag-of-words model is identified to be the optimum model, achieving validation accuracies of 88.6% and 78.0% for binary classification tasks on pore volume and $CO_2$ Henry's constant, respectively. Further, we developed multi-class classification models tailored to the probabilistic nature of quantum circuits, with average test accuracies of 92% and 80% across different classes for pore volume and $CO_2$ Henry's constant datasets. Finally, the performance of generating MOF with target properties showed accuracies of 97.75% for pore volume and 90% for $CO_2$ Henry's constant, respectively. Although our investigation covers only a fraction of the vast MOF search space, it marks a promising first step towards using quantum computing for materials design, offering a new perspective through which to explore the complex landscape of MOFs.

## Introduction

As the idea of quantum computers first arose in 1982 following Feynman's proposal[1], many theoretical and experimental works have explored the possibilities to accelerate scientific discovery using this technology. Recently, quantum machine-learning (QML) algorithms have surfaced as a promising alternative to their classical machine-learning approaches[2] to develop materials based on the advantages of quantum computers[3]. This shift is attributed to the algorithmic speed-up of quantum computers based on quantum parallelism and the non-classical correlations arising from quantum entanglement[3]. At the core of QML lies the exploitation of quantum bits (qubits), which unlike classical bits, can exist in a superposition of states and be entangled with one another. This allows quantum algorithms to perform complex calculations more efficiently based on quantum parallelism to explore multiple possibilities simultaneously[4]. Current quantum computing technology is categorized as being at the Noisy Intermediate-Scale Quantum (NISQ) stage, which refers to intermediate number of qubits (approximately 100 qubits), while lacking the capacity for large-scale error correction[3]. Consequently, the NISQ era poses significant challenges that prevent the demonstration of a clear quantum advantage over classical algorithms, due to the limited number of qubits and their susceptibility to errors[5].

As a result, the development of QML algorithms for NISQ devices largely focus on a hybrid quantum-classical approach with shallow-circuit depth, which delegates a part of the computation to a classical processor to manage qubit noise effectively[3, 6]. To this end, many researchers have suggested methods to apply QML in solving complex problems in materials science and chemistry by focusing on compatible QML methodologies. Specifically, Kanno et al. demonstrated that the physicochemical properties of the periodic materials can be successfully calculated with this concept by adopting a restricted Boltzmann machine-based variational quantum eigensolver (RBM-based VQE) for the band structure calculation of graphene[7]. This approach was further developed by Sajjan et al. as conduction bands of monolayer transition metal dichalcogenides, $MoS_2$ and $WS_2$, were simulated and filtered based on angular momentum symmetry[2]. Scientific endeavors to further extend the search space of materials with quantum computing led to the development of a QML algorithm for property prediction. Brown et al. introduced a quantum variational eigensolver-based circuit learning method to estimate the phases of high-entropy alloys (HEAs), with the model's performance nearing that of classical artificial neural network (ANN) models in classifying ternary and binary phases[8]. Naseri et al. explored the use of a supervised hybrid quantum-classical ML algorithm for the binary classification of $ABO_3$ perovskite structures, showcasing the applicability of QML in materials discovery[9]. Despite these advancements, previous studies have largely focused on relatively simple periodic systems such as transition metal dichalcogenides (TMDs) and perovskite structures, with challenges arising in scaling to larger systems like metal-organic frameworks (MOFs) due to the difficulty of individually encoding atoms into a constrained qubit resource.

To address this limitation, we adopted using quantum natural language processing (QNLP) as a novel approach for property-guided selection of MOF structures within a discrete design space efficiently. Before introducing the QNLP methodology in detail, it is important to emphasize that the quantum computing in the NISQ era should be regarded primarily as a conceptual and theoretical frontier, rather than a practical alternative that outperforms classical methods. As noted in the recent reviews, NISQ systems offer valuable opportunities to explore hybrid quantum-classical algorithms, but reliable demonstrations of quantum advantage remain limited due to noise and resource limitations[10]. The primary objective of this work is not to claim improvements in computational speedup or efficiency over classical machine learning techniques. Instead, we aim to establish a conceptual framework for applying QNLP to model the large molecular system, which go beyond the relatively simple periodic systems typically explored in previous quantum machine learning studies.

QNLP is an emerging field that combines quantum computing with natural language processing (NLP), aimed to utilize the principles of quantum mechanics to process and analyze natural language data more efficiently than classical computing methods. This concept was first introduced by Coecke et al.[11, 12] based on the idea that quantum computing is naturally suited to process high-dimensional tensor products and that NLP can utilize vector spaces to describe sentences[13]. In quantum computing, the state of a quantum system with multiple qubits is represented by a vector in a high-dimensional Hilbert space[14]. Since the operations or transformations applied to these qubits are represented by matrices, any quantum operations to the system of qubits are processed by performing tensor products of matrices to calculate the new state of the system[4]. This enables the representation of the complex, high-dimensional states and transformations fundamental to quantum computing. In the context of NLP, sentences can be represented in high-dimensional vector spaces, and the relationships between words involve complex transformations within these spaces. The potential of QNLP arises from its natural and efficient use of such calculations due to its inherent ability to operate in high-dimensional Hilbert spaces. As such, a hybrid quantum-classical algorithm based on QNLP was adopted from the previous studies for sentence generation[15] and music composition[16]. For sentence generation, Karamlou et al. performed a comparative study on two types of datasets: binary classification for identifying different sentence topics and multi-class classification for identifying news headlines. QNLP models were separately trained for these datasets, and these quantum models provided feedback to the classical model for sentence generation, resulting in 22 and 11 correct sentence generations out of 30 attempts for binary and multi-class classification models, respectively, in the classical simulator provided by the IBM Qiskit[15]. A similar quantum-classical framework was applied to music generation by Miranda et al[16]. The QNLP model was developed to identify types of music (between rhythmic and melodic), resulting in a test accuracy of 76% on real quantum hardware[16]. However, to the best of our knowledge, QNLP has yet to be applied to any molecular structures or materials.

In this work for the first time, we explored the possibility of using QNLP to model metal-organic frameworks (MOFs), which are crystalline porous materials consisting of metal clusters and organic ligands[17]. The choice of MOF as our target material was based on its modular nature, which is akin to how words form sentences. Their structural and functional characteristics can be designed through the rational selection of topologies and constituent building blocks, including organic ligands and metal-containing units[18-20]. This characteristic resembles how sentences with various subjects are formed based on the choice of words. This versatile architecture allows MOFs to be widely used in many applications such as gas sensors[21], drug delivery[22], gas storage[23], and separation[24]. However, the challenge in rationalizing component choices arises from the extensive search space required to select combinations of building blocks and topology. As the search space for MOFs continues to expand, addressing the curse of dimensionality becomes critical. This concern was previously emphasized by Brown et al. in the context of designing periodic materials with large degrees of freedom such as HEAs[8].

We adopted the hybrid quantum-classical method to construct a MOF generation framework with user-desired properties. Pore volume and $CO_2$ Henry's constant were chosen as target properties of interest as they reflect the physical and chemical properties of MOF structures, respectively. The former reflects a direct result of the MOF's structural characteristics, whereas the latter stems from the MOFs' ability to adsorb and interact with guest molecules, an attribute linked to the interplay of their geometrical and chemical properties[25], which amplifies the challenge of the task. By representing MOFs in terms of their constituent building blocks and topology, we simplified their qubit representation, thereby making quantum resources more accessible for the study of complex materials. The overall workflow of this work is summarized in Fig. 1. Initially, we describe the construction of MOF datasets, featuring a simplified representation of 450 MOF structures. These are categorized by target property classes, such as low, moderately low, moderately high, and high pore volume or $CO_2$ Henry's constant classes (Fig. 1a). Next, we rationalized our choice of the QNLP models based on the comparative study among four different approaches: the bag-of-words, DisCoCat (Distributional Compositional Categorical), and two sequence-based models (Fig. 1b). The multi-classification models were then developed, exploring ways to overcome the statistical limitations inherent to the QNLP model selected from the comparative study (Fig. 1c). These limitations stem from the probabilistic nature of the chosen QNLP model. As a result, QNLP models for categorizing target classes were developed with average test accuracies of 92% and 80% for pore volume and $CO_2$ Henry's constant datasets, respectively. Finally, these models were integrated with the classical MOF generation process, which generates text-based MOF input by randomly selecting topology and building blocks in each component's search space (Fig. 1d). The QNLP models functioned as answer sheets, providing feedback to the classical generation loop to navigate the search space of MOF topology and building blocks correctly. To test the performance of the overall process for generating MOF with target property, the model was tested on a classical Aer simulator by the IBM Qiskit[26] to generate MOFs with low, moderately low, moderately high, and high pore volumes and $CO_2$ Henry's constants. The average generation accuracies for MOFs with desired pore volume and $CO_2$ Henry's constant were 93.5% and 87%, respectively. In the last section, we discussed the extensibility of QNLP to other complex periodic materials based on 'sequence' as one of the promising parameters for representing chemical structures. Our study not only circumvents the qubit limitation but also opens new avenues for applying quantum machine learning in the exploration and design of advanced materials.

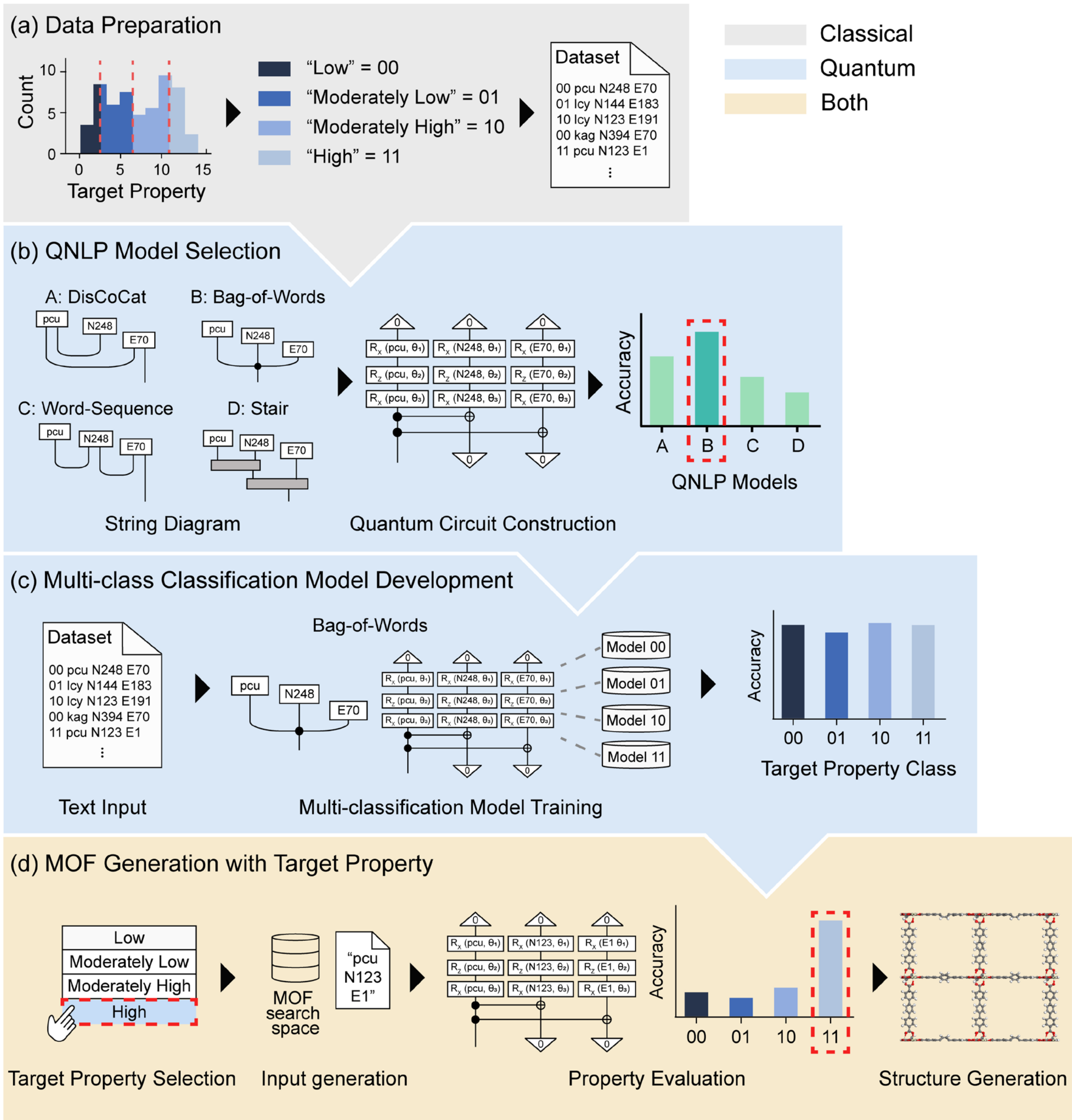

**Figure 1.** General overview of the study. Gray, blue, and yellow boxes indicate the classical computer-based, quantum computer-based, and hybrid quantum-classical computer-based processes, respectively. **(a)** MOF dataset construction from target property distributions encoded into four classes, and it is transformed into the text dataset consisting of the target class and its corresponding MOF name. **(b)** Comparative study among QNLP model candidates. Models A, B, C, and D refer to DisCoCat, bag-of-words (BoW), word-sequence, and stair models, respectively. **(c)** Multi-classification model development based on the choice of QNLP model from the QNLP model selection step. Four binary classification models by target property type (pore volume and $CO_2$ Henry's constant) are combined to complete a multi-classification framework. **(d)** MOF generation framework to build MOF structure with the desired target property.

# Results

## Building the MOF Dataset

The MOF dataset in this work consists of 3 topologies, 10 types of metal clusters (labeled as nodes in this work), and 15 types of organic ligands (labeled as edges in this work), with the number being small considering the current limitations in quantum resource availability (Fig. S1). Chemical molecules and materials possess complex degrees of freedom that influence their mechanical and chemical properties. This complexity arises because the individual components of chemical compounds are directly linked to their properties. For instance, MOFs that share the same types of metal nodes and ligand edges but differ in their topology (i.e., polymorphism) result in significant differences in surface area, which applies to the case of MIL-101[27], MIL-88[28], and MIL-53[29]. Furthermore, MOFs with identical topology and metal nodes but different organic ligands exhibit variations in chemical and mechanical properties, such as gas uptake and pore diameter[30]. This suggests that introducing a single type of topology or building block adds more than just a single degree of freedom, thereby complicating the selection of topology and building blocks given the manageable dataset size for the NISQ devices. Consequently, the dataset was restricted to include three topologies. This decision was based on the direct relationship between topology and our target properties, aiming to minimize the impact of variations in the building blocks, metal nodes, and organic ligands.

A pcu topology, one of the most fundamental topologies, was selected for its straightforward pore structure and high symmetry owing to a cubic arrangement[31]. The choice of kag and lcy was based on their structural distinction while maintaining the same node connectivity as the pcu topology, ensuring that the same building blocks can be interchangeably used (Fig. S1a). The MOF building blocks were chosen based on structural similarity in the molecular framework to ensure a broad and consistent distribution of target properties of pore volume and $CO_2$ Henry's constant. For the metal clusters, we considered structural analogs that form rod-like connections with the ligands. These clusters are composed of two metals covered by oxygen functional groups, represented by the formula, $\mathrm{M_2C_6}O_{12}X_6$ (where $\mathrm{M}$ denotes the metal and $\mathrm{X}$ represents the connection point with the ligand, as shown in Fig. S1b). For the organic ligands, cyclic compounds with various lengths ranging from 3.56 to 16.12 Å were considered (Fig. S1c). The datasets were evenly divided into four classes with regard to the aforementioned target properties: (1) low, (2) moderately low, (3) moderately high, and (4) high, as shown in Fig. 2a and c. The decision to use multiple target property classes rather than a simple binary classification (e.g. low and high) was made to enable the generation of MOFs within the more refined user-desired target property class. The class boundary was determined to ensure an even distribution among the classes, with 113 samples each for the low and high classes and 112 samples each for the moderately low and moderately high classes in terms of pore volume and $CO_2$ Henry's constant, thereby maintaining the uniformity of the dataset. Finally, low, moderately low, moderately high, and high classes were encoded into classical labels of 00, 01, 10, and 11, respectively, to directly compare with quantum models' prediction, as depicted in Fig. 1a. The next section will provide a detailed explanation of how the outcomes of the quantum model are processed and compared with the true labels. The MOF datasets for QNLP model training were then prepared with the target property class and its corresponding MOF name listed by topology, node, and edge names. For example, the MOF named 'pcu N248 E220' with low pore volume, was mapped into the label '00 pcu N248 E220' in our MOF dataset.

Fig. 2b and d show the influence of the MOF components (topology and building blocks) on the classes of pore volumes and $CO_2$ Henry's constants by showing the number of occurrences per topology or building block in each class. The counts can be considered as the numerical contribution of the individual building blocks for each class. Considering that 150, 45, and 30 MOF structures can be created per topology, node and edge type, respectively, the individual MOF components that appeared more than about 50% of possible structures in a class were considered 'class-significant' topologies or building blocks. This indicates that the MOFs with the corresponding components (topology or building block) are highly likely to possess that class property. Therefore, the 60-count is considered the boundary for class-significance for topology type and the 20-count is considered the boundary for class-significance for node and edge type. A class-significant component shows a skewed distribution, meaning it appears predominantly in a specific class rather than being evenly distributed across all classes. This indicates a strong association between that component and the corresponding property class. For example, a particular edge that appears primarily in the "moderately high" (10) class is likely to contribute to that property. Based on this assumption, the impact of the organic ligands on the pore volume dataset was greater than that of topologies and metal clusters, exhibiting class-significant edges in every class with clear skewed distribution favoring specific classes. Topology was the second most contributing factor, as class 11 consisted of 67 kag topology-based MOFs, whereas the metal clusters contributed the least, showing no class-significant building blocks. Edge contribution was the highest for the $CO_2$ Henry's constant dataset, although its distribution was broader than that of pore volume dataset. Unlike the pore volume dataset, the node distribution showed class-significance in classes 11 and 10, while no class-significant topology was observed. Understanding the distribution of class-significant building blocks is crucial, as it directly affects the difficulty of model training. For instance, generating MOFs with high pore volume (the 11-class) is considerably easier than generating MOFs with the moderately high pore volume (the 10-class) owing to the higher occurrence frequency of specific MOF components, such as E161, E229, and the kag topology in the pore volume dataset. Consequentially, there is a variation in the level of difficulty across the classes despite the even distribution of sample numbers, reflecting the continuous nature of MOF properties. The difference in class distribution between the pore volume and $CO_2$ Henry's constant datasets arises from the fact that Henry's constant primarily reflects adsorption strength, which is largely dictated by chemical interactions although pore geometry can have an indirect influence. This distinction is evident in their nonlinear correlation, as illustrated in Fig. S2.

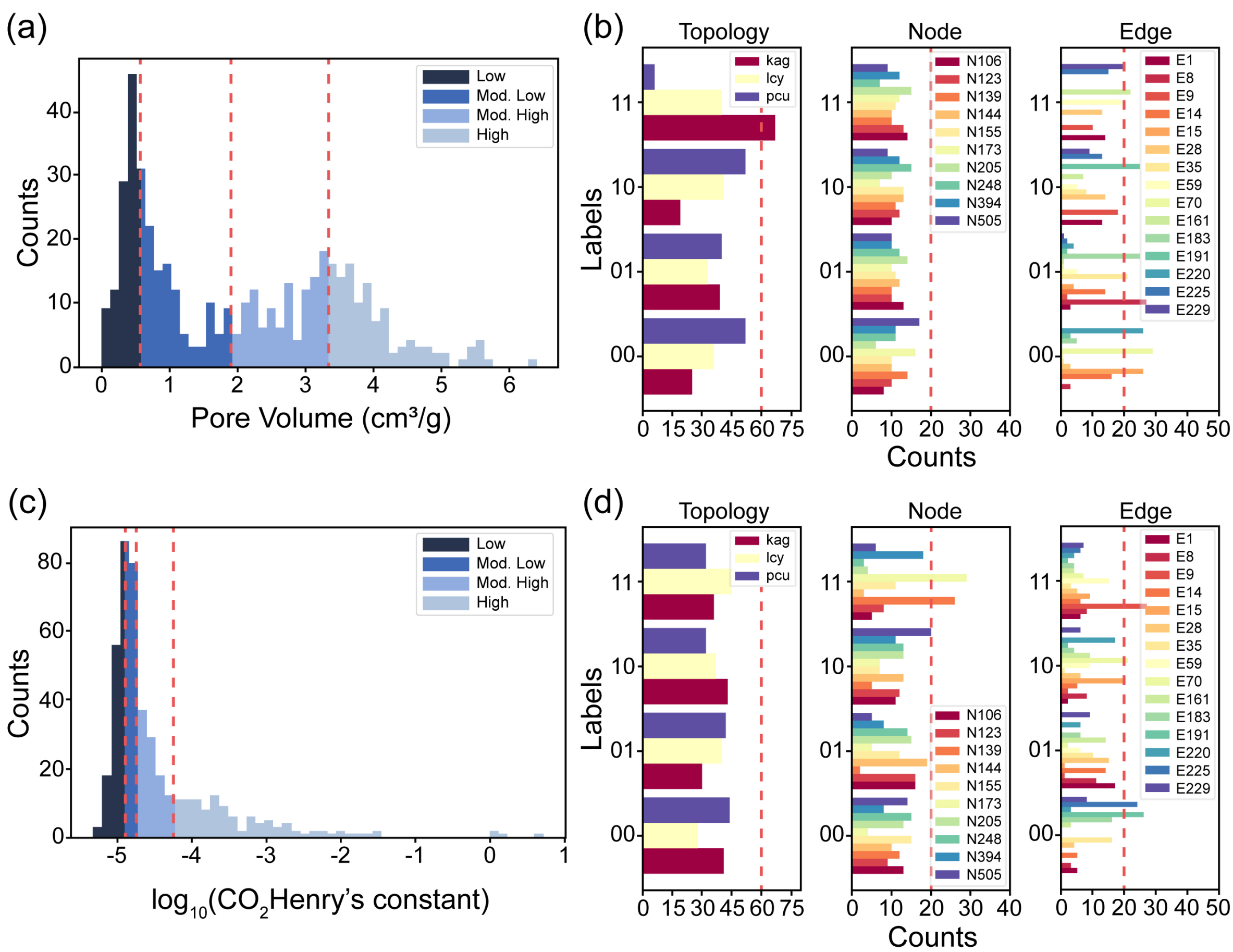

**Figure 2.** Distributions of 450 hypothetical MOFs based on **(a)** pore volumes and **(c)** the logarithm (base 10) of $CO_2$ Henry's constants (The unit of $CO_2$ Henry's constants is mol/kg/Pa). The datasets were divided into four classes based on 25th, 50th, and 75th percentiles of the pore volume and $CO_2$ Henry's constant values. Vertical dotted lines indicate corresponding class boundaries. Counts of classes per topology type, node (metal clusters) type, and edge (organic ligands) type based on **(b)** pore volumes and **(d)** $CO_2$ Henry's constants. Labels 00, 01, 10, and 11 represent low, moderately low, moderately high, and high pore volumes and $CO_2$ Henry's constants, respectively. Vertical red dotted lines represent a count boundary of the class-significance.

## Mapping the MOF Dataset onto Different QNLP Models

The selection of existing QNLP models is motivated by two key considerations. First, the modular structure of the MOF dataset used in this study exhibits a resemblance to the compositional nature of simple sentence datasets. Each MOF in our dataset is constructed from three discrete components, topology, node, and edge, similar to how basic sentences are composed of subjects, verbs, and objects. Since our dataset was intentionally limited in complexity to reduce variation in structural degrees of freedom, the property classification task resembles a semantic classification problem in natural language processing, where the presence of specific components rather than their sequence or grammatical relationship plays a dominant role. For instance, Lorenz et al. compared QNLP models such as bag-of-words, DisCoCat, and sequence-based models in classifying simple sentences with different subject matter and clause types. Their results indicated that bag-of-words performed best in tasks where classification depended mainly on word choice, rather than syntactic structure. Likewise, our MOF property labels correlate with the occurrence of specific edges and topologies, as shown in Fig. 2(b) and (d), suggesting that component-wise features are influential. Second, there is historical precedent in applying classical NLP models originally designed for human language to molecular and materials data. Examples include the use of SMILES string, MOFkey representations[32], and other domain-specific textual encodings to predict molecular properties, generate new molecules and learn structure-function relationships[33]. As summarized by Öztürk et al., these approaches treat chemical structures as "unstructured languages" in which the sequential or co-occurrence patterns of tokens encode chemically meaningful information[34]. Since our work is the first to apply quantum natural language processing (QNLP) to materials design, we believe that adopting well-established QNLP models is a scientifically grounded starting point.

For quantum model training, the aforementioned MOF dataset needs to be transformed into quantum circuits, which requires pre-processing of the dataset based on the choice of QNLP models. In order to implement a quantum model for the property classification, four QNLP models were explored: BoW model, DisCoCat model, and the two word-sequence models. The MOF representation of topology, node, and edge was mapped into a high-dimensional vector space where each dimension corresponds to a unique component of the MOF. The method by which these components are processed, akin to grammatical structure, can be mathematically depicted through tensor products, with these operations on the component vectors that vary distinctively across different QNLP model types. The graphical representations of these complex vector spaces and the mathematical functions connecting them are known as string diagrams[35] as illustrated in Fig. 1b and Fig. 3a. Based on these diagrams and the choice of ansatz, the quantum circuits can be constructed as MOF component vectors are mapped into individual qubits, and quantum gate operations implement the tensor products. The ansatz is a proposed structure of quantum circuit consisting of adjustable circuit parameters such as the number of qubits or unitary operations within a given vector space. The qubits are initially in the $|0\rangle$ states and then evolve through a sequence of unitary operations (i.e. gates). Based on the choice of the IQP (Instantaneous Quantum Polynomial) ansatz, the unitary operations employed were Hadamard gate (H) and rotation gates of $R_x(\theta)$ and $R_z(\theta)$, which rotate a qubits state around the respective axis by an angle, $\theta$. Readers might refer to the method section for further explanation about the choice of ansatz.

In this paper, we will focus on the BoW model to specifically explain how the QNLP model transforms classical data into a quantum circuit and identify which components of the quantum circuits are optimized against the true labels. For information on other QNLP models, readers may refer to Note S1. The BoW model, based on a Frobenius algebra, transforms text into a sparse vector of word counts and represents the output as the component-wise multiplication of these vectors, $s$[35]. That is, each MOF component vector (topology, node, and edge) is mapped to the high-dimensional vector space obtained from the component-wise multiplication, independent of word order. In the associated string diagram, each component vector is represented as a separate box and the component-wise multiplication is represented as a 'merge-dot' as shown in Fig. 3a. The focus of this model would be on the presence or absence of specific component vectors over the explicit connections among them to indicate grammatical structure or sequence in other models such as DisCoCat and the word-sequence models (Note S1).

The transformation into a quantum circuit involves mapping individual MOF component vectors onto separate qubits, with unitary operations then used to manipulate the overall quantum state of these qubits (Fig.

3b). The number of qubits per component vector varies depending on the task, and this will be discussed in the next section. Initially, all qubits representing corresponding MOF components are in a default state, $|0\rangle$ (Fig. 3b(i)). The quantum state of each qubit is then uniquely altered through a sequence of unitary operations (Fig. 3b(ii)), and the rotational angles of these operations, $\theta_{c.n}$ (where $c$ and $n$ represent MOF component and integer, respectively), are quantum circuit parameters optimized through quantum machine learning. The merge-dot of the string diagram is encoded as the controlled-not (CNOT) gates (Fig. 3b(iii)), which naturally encode component-wise relationships between MOF components at the circuit level. These relationships refer to the logical conjunctions that determine whether particular combinations of topology, metal node, and organic edge co-occur in a MOF instance. The CNOT gate performs a conditional operation where the state of a target qubit flips only if the control qubit is in the $|1\rangle$ state, thereby entangling qubits and capturing the joint presence of specific components. This reflects the Frobenius algebra operation in the classical BoW model, where the value of the output is influenced only when both contributing components are present. Through these entangling operations, the circuit encodes the combined influence of MOF components on the overall quantum state, allowing it to capture their interactions without the need for explicit tensor product expansions or matrix multiplications, which are typically required in classical NLP models[36]. These interactions are embedded in the evolved quantum state and can be statistically inferred from repeated measurements of the quantum circuit. Finally, the predicted class label is determined by measuring output qubit from the quantum circuit. This qubit, associated with the predicted class label and delivering the outcome of the circuit measurement, is referred to as an ‘open’ wire qubit, as it is represented by the wire coming out of the component in the quantum circuit[35] (Fig. 3b). For instance, in a binary classification task for pore volume as an example, ‘low’ and ‘high’ pore volume classes can be encoded as the states $|0\rangle$ and $|1\rangle$ of the open wire qubit, respectively.

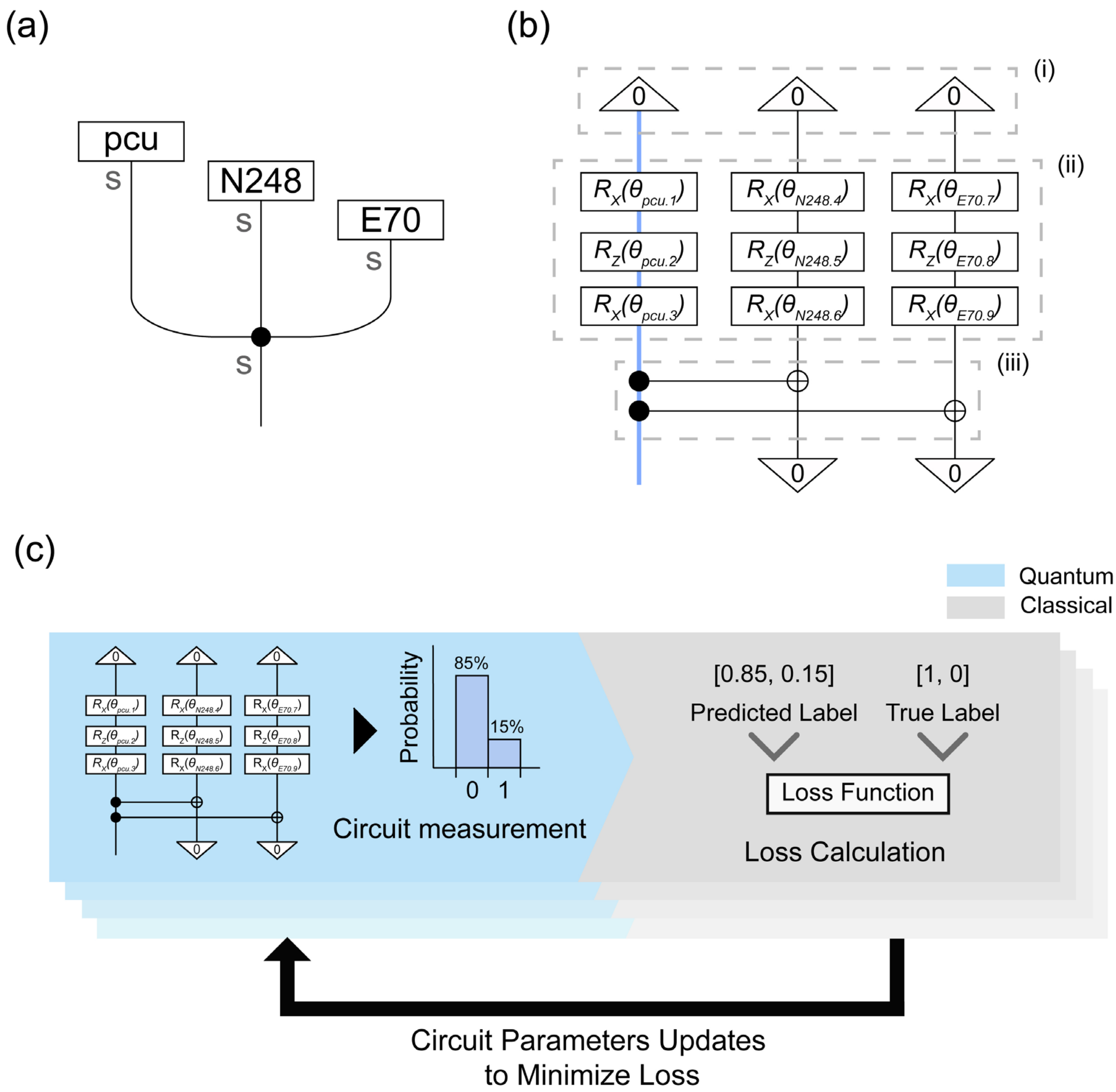

**Figure 3. (a)** String diagrams of MOF representation based on BoW model, and **(b)** its corresponding quantum circuit with a single-qubit configuration for binary classification based on IQP ansatz. Dotted line boxes represent **(i)** initial quantum states, **(ii)** circuit parameters and **(iii)** controlled-not gates which indicating merge-dot. Open-wire that carries predicted class label is highlighted as blue line. **(c)** An overall process of quantum machine learning.

Once the quantum circuit is constructed based on the choice of QNLP model, the next step is to introduce a quantum machine learning algorithm to optimize the circuit parameters. This optimization ensures that the outcomes from circuit measurements converge to the desired results (i.e. true label, Fig. 3c). The output of the quantum circuit is obtained by repeatedly measuring the circuit a specific number of times, known as shots. For instance, in binary classification of pore volume, the measurement outcome is either 0 or 1, where 0 and 1 represent low and high pore volume classes, respectively. These outcomes can be represented as a probability distribution of 0s and 1s as the number of measurements (i.e. shots) accumulates. Initially, the circuit parameters are set at random angles of $\theta_{c.n} \in [0, 1]$ resulting in the probability distribution that leads to random quantum states. These parameters are fixed based on the component type, although some parameters are shared between quantum circuits corresponding to different MOFs when identical components appear in multiple MOF structures. In total, 84 parameters are used (3 circuit parameters per MOF component which consists of 3 topologies, 10 nodes, and 15 edges) for this binary classification task example, and are uniquely combined to represent each MOF. For instance, the circuit parameters for a qubit representing the 'pcu' topology are the same across all MOF structures (e.g., pcu N173 E220, pcu N505 E229, etc.).

This parameter-sharing scheme has both technical and chemical implications. From a technical perspective, it significantly reduces the number of trainable parameters and simplifies quantum circuits, making implementation feasible within current NISQ hardware constraints. All state-of-the-art QNLP models implemented via lambeq[37], including those examined in this study (BoW, DisCoCat, and sequence models), employ this strategy due to the technical constraints and resource limitations inherent in current quantum computing platforms. However, such scheme makes an important chemical assumption that identical building blocks contribute consistently to MOF properties regardless of their local chemical environment. In reality, the same organic linker may exhibit different electrical properties[38] or coordination modes[39] when paired with different metal nodes or integrated into different topological frameworks. This parameter-sharing scheme therefore represents a trade-off between computational feasibility and chemical accuracy, where we prioritize demonstrating QNLP methodology over capturing the full complexity of structure-property relationships in reticular chemistry. While this limitation constrains the chemical realism of our current approach, it reflects the necessary simplification required for proof-of concept demonstrations within current quantum hardware capabilities rather than a fundamental limitation of QNLP approaches.

The initial measurement outcome, based on random circuit parameters, often deviate from the desired results, such as 1, assuming the MOF has a high pore volume. Thus, these parameters of the QNLP model (i.e. model weights) are initially set and then iteratively adjusted during training based on stochastic approximation of the gradient derived from the loss function[35] (Fig. 3c). This involves a vectorization of the circuit measurement outcomes to match the dimension of the true label, which is represented in one-hot encoding. For instance, binary quantum states $|0\rangle$ and $|1\rangle$, occurring with 85 and 15 shots out of a total of 100 (i.e. probabilities of 85% and 15%), are transformed into the vector $[0.85, 0.15]$, and based on which the loss is then calculated. The iterative process, which involves calculating the loss and updating the model's weights, is performed by a classical computer while the measurement of the predicted label is carried out by quantum computing method. Unlike classical NLP models, which require explicit probability normalization through activation functions like softmax, QNLP inherently generates probability distributions directly from quantum measurements. This eliminates additional computational overhead associated with feature aggregation and decision-making in classical models, offering a more streamlined and computationally efficient approach[36]. The difference in quantum machine learning over their classical counterpart is that the parameters being optimized, $\theta_{c.n}$, are circuit parameters that directly influence the transformation of quantum states, thereby manipulating the probability distribution of the measurement outcomes.

## Comparative Study on QNLP Models

The comparative study was initially conducted to identify the most suitable compositional model for our MOF dataset. For this experiment, the dataset was prepared for the binary classification task (as opposed to multi-class classification task) due to the task complexity and higher computational cost required for the multi-class classification. Therefore, the pore volume of 1.05 $cm^3/g$ and $CO_2$ Henry's constant of $1.819\times10^{-5}$ mol/kg/Pa were set as the class boundaries for categorizing MOFs with either low or high pore volumes and $CO_2$ Henry's constants, respectively. The target classes of low and high were encoded as labels 0 and 1, respectively, to be directly compared with the quantum states (i.e. predicted label) from the circuit measurements. The quantum circuits of MOF were constructed by pre-processing the MOF dataset based on each candidate QNLP models (corresponding to Fig. 3b, and S3d, e, f). These circuits were trained to associate MOF names accurately with their corresponding property labels. As mentioned in the previous section, this is achieved by iteratively refining their circuit parameters, which are associated with quantum states to minimize classification errors, as the quantum circuits initially result in random quantum states. For sequence-sensitive models (DisCoCat, word-sequence with Cups, and word-sequence with Stairs), optimal component orderings were determined by evaluating all possible permutations of MOF component sequences: topology (T), node (N), and edge (E). This generated six unique sequences, ENT, ETN, NET, NTE, TEN, and TNE. Readers are encouraged to refer to the method section and Note S2 for detailed discussion of dataset preparation and the sequence optimization process.

As illustrated in Fig. 4, all models exhibited a converging trend over the epochs characterized by decreasing loss, which was subsequently followed by a flattening pattern. For the pore volume dataset (Fig. 4a), both word-sequence models showed relatively inferior performance. The word-sequence with Cups model (Cups model) converged at accuracies of 0.693 and 0.69 and loss of 0.623 and 0.88, and the word-sequence with Stairs model (Stairs model) converged at accuracies of 0.86 and 0.88 and loss of 0.419 and 0.392 for training and validation dataset, respectively. The inferior training performance of sequence-based models was also observed in the $CO_2$ Henry's constant dataset (Fig. 4b). The performance gap between training and validation further increased as losses converged at 0.761 and 1.001 for the Cups model and 0.687 and 0.713 for the Stairs model. Surprisingly, the DisCoCat model showed relatively good convergence behavior, with loss minimized below 1 for both datasets. These outcomes aptly illustrate that MOFs are not sequence-dependent, but rather, they are the aggregate sum of each structural component. Although both DisCoCat and word-sequence models are syntax-sensitive approaches, the rule respecting text order is unsuited to MOF structures. This is because the criterion for completing a MOF structure depends on whether it constitutes a complete set of topology and building blocks rather than the order in which they are listed. This explains the discrepancy in performance between the DisCoCat model and the word-sequence models. The pregroup grammar of DisCoCat is rigidly governed by the syntactic arrangement of the words[13], whereas the word-sequence models are solely driven by the left-to-right sequence, irrespective of the sentence's components[35].

The best-performing model for our MOF datasets was the bag-of-word model, which illustrates that the modular structure of the MOF dataset is well suited for component-wise multiplication of parsed MOF tensors rather than sequence-based data processing (Fig. 4, Note S2). A similar result was previously reported by Lorenz et al. as the DisCoCat and the BoW models showed comparable performance, while the word-sequence model showed inferior performance for a meaning classification task involving two distinct subjects[35]. Here, the BoW model converged at the highest accuracies among models as 0.925 and 0.93 for the pore volume training and validation dataset, with losses at 0.169 and 0.149. The training and validation accuracies for the $CO_2$ Henry's constant dataset decreased to 0.807 and 0.78, with corresponding losses of 0.443 and 0.523. This lower performance can be attributed to the greater complexity of Henry's constant compared to pore volume, as it is influenced by both chemical interactions and geometric factors, making it more challenging for the model to learn and predict accurately (this was illustrated by the broader distribution of MOF components in Fig. 2d compared to Fig. 2b). Furthermore, the QNLP-based BoW model achieved comparable validation accuracy and even higher test accuracy than traditional machine learning models, including Random Forest and Support Vector Machine (SVM), despite relying on fewer trainable elements and a fundamentally different computational paradigm (Table S2). This indicates that QNLP offers strong generalization capabilities under the constraints of current quantum hardware. Readers are encouraged to refer Note S3 for further details. Therefore, the BoW model was chosen as the best candidate to develop the multi-class classification model for pore volumes and $CO_2$ Henry's constants.

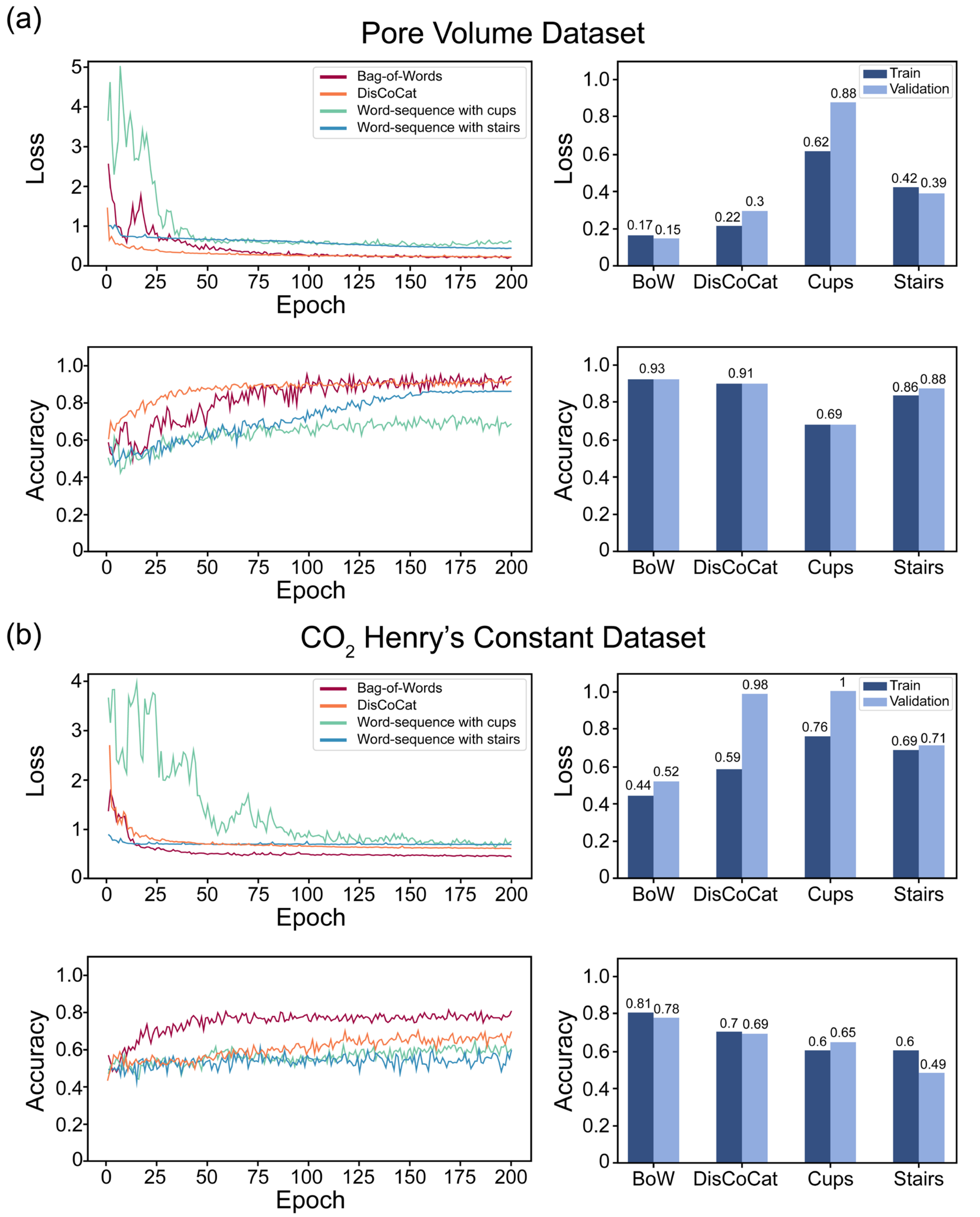

**Figure 4.** Training dynamics and final accuracies and losses for four QNLP models (the bag-of-words (BoW), DisCoCat, word-sequence with Cups (Cups), and word-sequence model with stairs (Stairs) models) trained on **(a)** pore volume and **(b)** $CO_2$ Henry's constant dataset for binary classification task. The left-hand panels display the training losses and accuracies as a function of epochs, showing convergence behavior. The right-hand panels summarize the final training and validation looses (top) and accuracies (bottom) for each model.

## Development of Multi-Class Classification Models based on the Probabilistic Nature of Quantum Machine Learning

In the previous section, MOF structures were assigned 0 or 1 output values based on low/high pore volume/$CO_2$ Henry's constants. However, in practice, we want more refined categories for classifications (e.g. very high pore volume, high pore volume, average pore volume, low pore volume), especially for inverse design purposes. Thus, the transition from binary classification to multi-class classification is important for practical purposes. Accordingly, the MOF datasets divided into four classes (low/moderately low/moderately high/high) were transformed into quantum circuits for multi-classification applications based on the choice of data representation schemes, the BoW model from the previous section (Fig. 5a). The fundamental difference between the circuit construction for binary and multi-classification is the number of qubits required to represent the MOFs, which leads to a difference in the probability of obtaining meaningful outcomes. The output of the quantum circuit is delivered by a specific number of qubits. For example, a quantum circuit of MOF representation for binary classification requires a single output qubit to deliver the probability distribution of predicted labels of 0 and 1 while that for the multi-classification requires two output qubits to represent the probability distribution of predicted labels of 00, 01, 10, and 11. Thus, the QNLP model involves a post-processing step known as post-selection, which retains only the measurement outcomes related to the predicted class label and filters out the remaining measurements based on a specific quantum state, 0-effect[35]. The term, 0-effect, is used as it determines whether to keep or discard the measurement outcomes based on whether the state of the post-selected qubits matches the computational basis state, $|0\rangle$. For instance, in a binary classification task implemented with a quantum circuit using three qubits to encode possible classes, post-selection is used to consider only those states where both second and third qubits are in the $|0\rangle$ state, such as $|000\rangle$ and $|100\rangle$. This selective measurement filters out results where the post-selected qubits are $|1\rangle$ (i.e., states $|011\rangle$ or $|110\rangle$), thus focusing the analysis on the relevant subset of the quantum state space. That is, the post-selection is to measure all quantum states but only interpret the state of the open-wire qubits for final classification, considering the states of the other qubits as ancillary information that supports the coherence of the quantum states but isn't directly used for decision-making.

In the MOF representation for binary classification, as discussed previously, each MOF is assigned a binary label 0 (i.e. low) or 1 (i.e. high) depending on their pore volume or $CO_2$ Henry's constant properties. The assigned label becomes the true label compared with the predicted label as a result of the circuit measurement. This requires the post-selection on $|0\rangle$ state of the two qubits to retain meaningful quantum states ($|000\rangle$ and $|100\rangle$) from the total probability distribution resulting from the circuit measurements. For example, the quantum circuit based on BoW model for binary classification allocates a single qubit for each MOF component vector (i.e., topology, node, and edge). This results in a total of 3 qubits, allowing $2^3$ possible quantum states, from 000 to 111 (Fig. 5b). Since the measurement outcome needs to be converged into single qubit states of 0 and 1, the measurement outcomes of 000 and 100 are interpreted as the probability distribution of predicted labels of 0 and 1, respectively. Based on the post-selection criterion, the remaining measurement outcomes are discarded. This indicates that assuming all possible outcomes have a uniform distribution (e.g. all $2^3$ quantum states are observed 125 shots each when measuring the circuit 1000 shots), only two in $2^3$ shots would result in a meaningful outcome representing the probability distribution of predicted labels (Fig. 5b). The prediction accuracy (probability distribution) and loss are then calculated based on these post-selected outcomes. In this case, the measurement produces counts, $c(b_2 b_1 b_0)$, with $\mathrm{b_j} \in \{0, 1\}$ across all three-bit outcomes. With $\mathrm{S}$ total circuit executions (i.e. shots), the kept shots after post-selection, $K_{bi}$, are

$$K_{bi} = c(000) + c(100) \qquad (1)$$

The post-selected class probabilities used for training and evaluation are then:

$$p_0 = \frac{c(000)}{K_{bi}}, \qquad p_1 = \frac{c(100)}{K_{bi}} \qquad (2)$$

where $p_0 + p_1 = 1$. Here, $p_0$ and $p_1$ are the predicted probabilities for class 0 and class 1, and $K_{bi}/\mathrm{S}$ is the empirical post-selection success rate. These probabilities are used to compute the binary cross-entropy loss, $L_{BCE}$, which trains the model by penalizing divergence from the ground-truth labels. With batch size $N$, the loss is:

$$L_{BCE} = -\frac{1}{N}\sum_{i=1}^{N}[y_0^{(i)} \log(p_0{}^{(i)}) + y_1^{(i)} \log(p_1{}^{(i)})] \qquad (3)$$

Where $\mathrm{y}^{(i)} = [y_0^{(i)}, y_1^{(i)}]$ is the one-hot ground-trough label for sample $i$ and $[p_0^{(i)}, p_1^{(i)}]$ is the corresponding predicted probability vector after the post-selection normalization.

The MOF representation for the multi-class classification, on the other hand, further increases the number of post-selections as it necessitates at least two open-wire qubits to represent the assigned label. Four classes of low, moderately low, moderately high, and high are encoded into two-qubit configurations of 00, 01, 10, and 11, and then compared with the resulting probability distribution. As mentioned in the previous section, the output of the BoW model is represented as a n-dimensional single tensor, derived from the component-wise multiplication of n component vectors. For instance, $n = 3$ to represent a single MOF for binary classification, as it consists of three components of topology, node, and edge. This component-wise multiplication is encoded into the quantum circuit by entangling individual component qubits with an open wire qubit using controlled-not operations, representing the merge-dot of the string diagram (Fig. 3b(iii)). Thus, for the multi-classification task, two open wire qubits are required to deliver quaternary quantum states, necessitating the duplication of required qubits by $n$ (Fig. 5a). This enables the comparison with true labels encoded as a two-qubit configuration but further increases statistical limitations by requiring the post-selecting of 4 qubits (Fig. 5c). Here, the measurement returns counts, $c(b_5b_4b_3b_2b_1b_0)$, with $b_j \in \{0, 1\}$ across all six-bit outcomes. Since the post-selection retains only outcomes where the ancillary qubits are measured in the $|0\rangle$ state, the kept shots after post-selection for multi classification, $K_{multi}$, are:

$$K_{multi} = c(000000) + c(010000) + c(100000) + c(110000) \qquad (4)$$

The post-selected class probabilities are then normalized as:

$$p_{00} = \frac{c(000000)}{K_{multi}}, \quad p_{01} = \frac{c(010000)}{K_{multi}}, \quad p_{10} = \frac{c(100000)}{K_{multi}}, \quad p_{11} = \frac{c(110000)}{K_{multi}} \qquad (5)$$

where $p_{00} + p_{01} + p_{10} + p_{11} = 1$, ensuring proper probability normalization after post-selection. For multi-class classification, the cross-entropy loss $L_{CE}$ is computed as:

$$L_{CE} = -\frac{1}{N}\sum_{i=1}^{N}\sum_{j=1}^{4} y_j^{(i)} \log\left(p_j{}^{(i)}\right) \qquad (6)$$

Where $j = \{00, 01, 10, 11\}$ represents the four classes, $y_j^{(i)}$ is the one-hot encoded ground-truth label for class $j$ for sample $i$, and $p_j{}^{(i)}$ is the corresponding predicted probability after the post-selection normalization. That is, four in $2^6$ shots would yield meaningful results, assuming the uniform distribution of total outcomes, and this notably reduces statistical significance by a factor of 4 compared to the binary classification situation. The number of post-selections required grows exponentially as the number of classes increases. This indicates that the number of circuit measurements also needs to increase exponentially in accordance with the number of post-selections to obtain meaningful statistics. To address this issue, the number of circuit executions for binary and multi-class classification models discussed in this section is set at 2048 and 8192 shots, respectively. One might also notice that the circuit designs for multi-class classification (Fig. 5a) involve fewer trainable parameters than those for binary classification (Fig. 3b). This difference arises from how the lambeq software internally handles circuit generation. Specifically, lambeq restricts the number of trainable gates in multi-qubit configurations to enforce well-defined entanglement patterns and avoid over-parameterization. We encourage readers to refer to the Methods section and Note S4 in the Supplementary Information for a detailed explanation for this constraint.

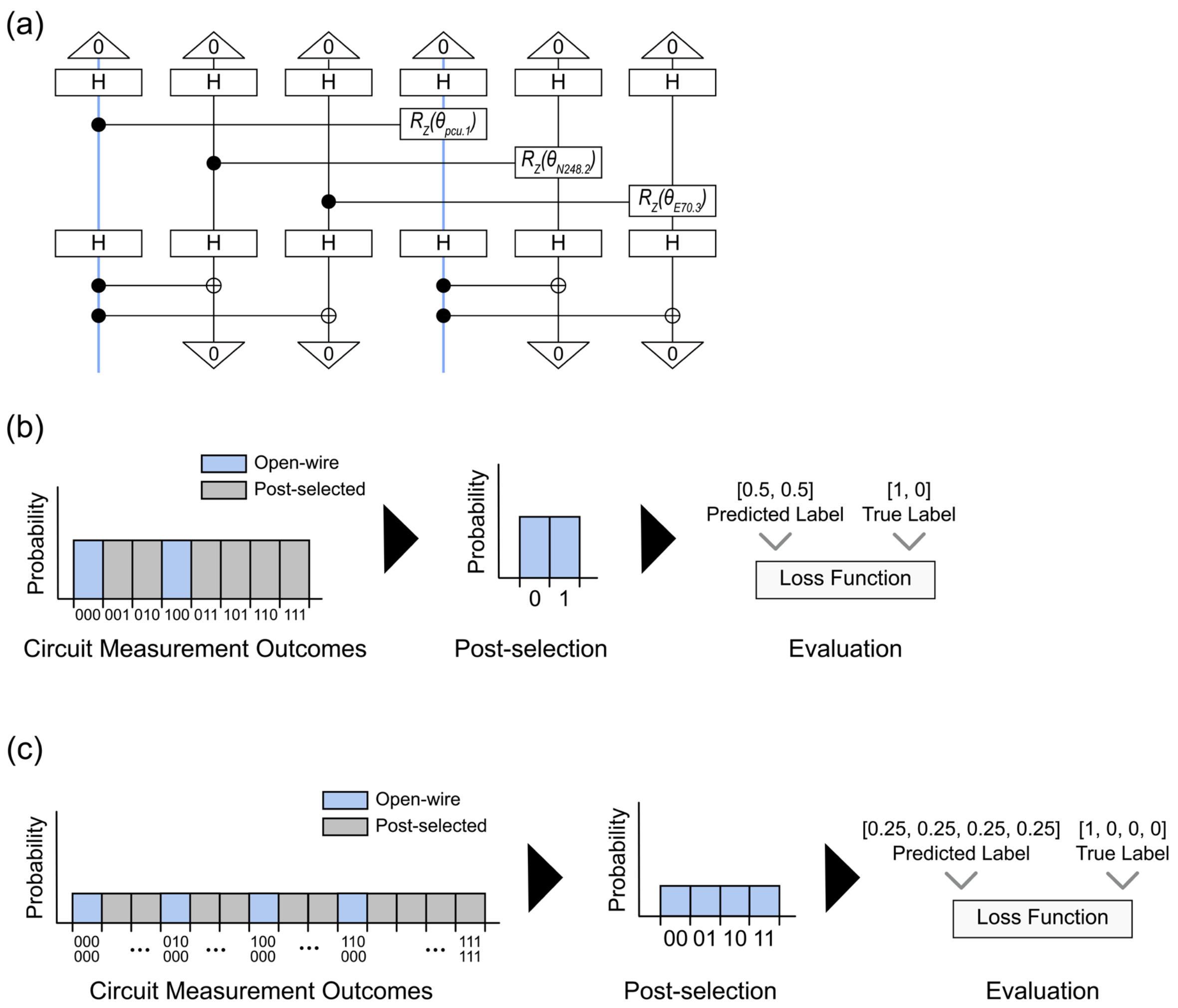

**Figure 5. (a)** Quantum circuits of MOF named pcu N248 E70 with two-qubit configuration for multi-classification tasks. Note that the blue lines represent open wires, and triangles pointing downward represent

qubits with a 0-effect. **(b)** A post-selection process assuming the quantum circuit measurement result in a uniform distribution for binary classification and **(c)** multi-classification tasks. The circuit measurement outcomes after the post-selection is vectorized to be matched with one-hot encoding form of the true labels.

In addition to the difference in qubit requirements, the limited sample size per class makes model training for multi-class datasets difficult. Splitting the datasets with binary labels results in 225 samples per class, while the datasets with multi-class labels result in 112 or 113 samples per class out of a total of 450 samples. Therefore, the task complexity of training models for datasets with multi-class labels is much higher than that with binary labels as illustrated in Fig. S5. One way to circumvent the limitation without increasing the overall dataset size or circuit depth is to transform the multi-class datasets into binary label-like datasets by separating them in a 20:80 ratio. For example, BoW model can be trained for binary classification based on the dataset divided into 113 samples of 'low' class with label 0 and the rest of the 337 samples labeled with 1. Similarly, 112 samples of 'moderately low' class would be labeled as 0, and the rest of the 338 samples would be labeled as 1; therefore, the trained BoW model becomes a 01-specialized model. Based on this strategy, the required qubits per MOF data point are maintained at three qubits; therefore, the number of circuit executions to obtain meaningful results for each class can be kept as 25 % of implementing the single multi-classification model. Fig. 6 illustrates training results of the four BoW models based on 00, 01, 10, and 11-specialized datasets. The training history showed good convergence behavior with smooth exponential decay, similar to that of the 50:50 binary dataset (Fig. 4), while the multi-class datasets showed large fluctuations over epochs (Fig. 6a, b). The prediction accuracies showed consistent results with significant difference between multi-class and class-specialized datasets (Fig. 6c, d). The single multi-classification model showed inferior performance over individual binary classification models trained on each class-specialized dataset by showing the average differences of 40.6 % and 36.7% for the pore volume and $CO_2$ Henry's constant, respectively. Consequentially, these results reflected increased task complexity in the multi-class datasets, which were caused by changes in trainable sample size.

Among BoW models for binary classification, a unique U-shaped trend was observed in the accuracy for both pore volume and $CO_2$ Henry's constant as the values started at about 0.9 in 00, reached the minimum value in 01 or 10, and then recovered in 11-specialized datasets. This is mainly because the overall trend of the building block dependency is focused on the low and high-class datasets, and the moderately low-class datasets have the least number of class-significant building blocks (Fig. 2b, d). Comprehensive evaluation metrics, validate this relationship that the $Bi_{00}$ and $Bi_{11}$ models achieve precision and recall above 0.74 with F1 scores reaching 0.77, while $Bi_{01}$ and $Bi_{10}$ models show reduced performance, consistent with their fewer distinguishing structure features (Table S3). Importantly, all class-specialized binary models substantially outperform the single-multiclass mode (Fig. 6c, b), which struggles with F1 scores of only 0.4 - 0.42 and exhibits systematic misclassification between adjacent classes (Fig. S10, Fig. S11). Despite the trade-off of training multiple models, the author believes this approach is a reasonable solution to maintaining dataset size and the complexity of the quantum model, considering the quantum resource in the NISQ device. Thus, this multi-QNLP model-based architecture was adopted to further experiment with the MOF generation task.

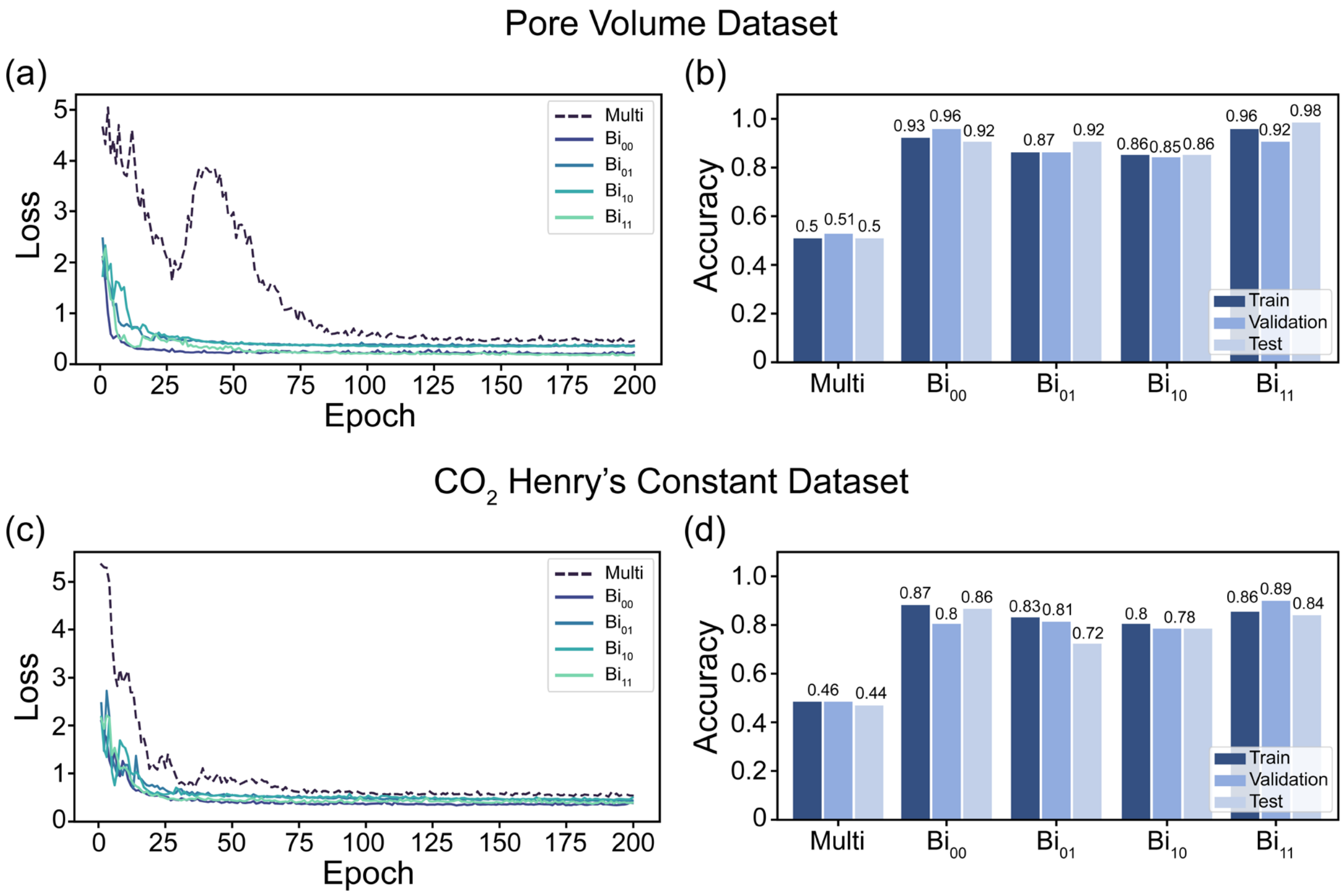

**Figure 6.** Training history of loss based on **(a)** pore volume and **(b)** $CO_2$ Henry's constant dataset. Accuracy bar graphs of **(c)** pore volume and **(d)** $CO_2$ Henry's constant datasets. Note that Multi, $Bi_{00}$, $Bi_{01}$, $Bi_{10}$, and $Bi_{11}$ represent BoW models trained based on the multi-classification dataset, 00-specialized dataset (binary classification dataset divided into classes 00 and 01+10+11), 01, 10, and 11-specialized datasets, respectively.

## Inverse Design of MOFs with Target Property

As sentences are considered logically correct based on word choice and grammar, MOFs can be considered complete only when it satisfies the essential components and follows a consistent set of structural rules, referred to there as the MOF grammar. A valid MOF must include three core components, topology, metal node, and organic linker (edge). In this study, we devised a classical MOF text input generation loop that randomly samples MOF components and assembles them into syntactically valid text representations based on defined MOF grammar rules, which are then evaluated by QNLP classifiers trained on pore volume and $CO_2$ Henry's constant datasets (Fig. 7a). The MOF grammar used in our framework consists of two key constraints. First, each MOF must follow a fixed component order, topology, followed by node, and then edge, which is consistent with the structure of the original dataset. Second, each MOF must include only one of each component type, meaning that mixed-metal or mixed-linker combinations (e.g. "pcu N123 N106 E70") are not allowed. These rules ensure that all generated MOF inputs are interpretable by the QNLP classifiers and structurally coherent with the dataset.

Once the text-based MOF input data is generated, it is transformed into the quantum circuit based on the BoW string diagram. Then, the quantum classification models discussed in the previous section evaluate the input MOF's property based on the probability distribution obtained from the circuit measurements (Fig. 7b). A single MOF quantum circuit is used as an input for four distinct classification models: Model 00 (i.e. BoW models trained on datasets specialized for label '00'), Model 01, Model 10, and Model 11. The interpretation of total probability distributions from these four models is based on the relative probability distributions against a class label '0', as each model is specialized to recognize a specific range of properties. Here, the label '0' represents specialized property classes (i.e., low, moderately low, moderately high, and high) which corresponds to quantum state '0' from the classification models. For instance, if the '0' state (i.e. specialized property class) of the MOF quantum circuit is measured with probabilities of 5%, 1%, 3% and 95% in the $CO_2$ Henry's constant models 00, 01, 10, and 11, respectively, the MOF is likely to have a high $CO_2$ Henry's constant with 91.3% relative probability (Fig. 7b). Finally, the MOF is evaluated to determine if it possesses the desired property, based on whether the highest relative probability exceeds a set threshold. The threshold cutoffs for each target property were determined through comparative calculations, where various threshold values were tested to identify the optimal threshold value that maximizes the inverse design performance (Note S5, Table S4). Based on the calculation, the threshold was set to 85% for the pore volume dataset and 65% for the $CO_2$

Henry's constant dataset. The MOF that meets the threshold goes through the structure generation process, while those that do not are redirected back to the input generation stage. This cycle continues until a generated MOF satisfies the threshold condition. The structure generation process based on the MOF text input is done by classical computation based on our in-house software, PORMAKE[40] (Fig. 7c). The MOF name's raw text generated in step 1 (e.g., pcu N123 E9) can be directly inputted, as it adheres to the naming convention of building blocks used in PORMAKE.

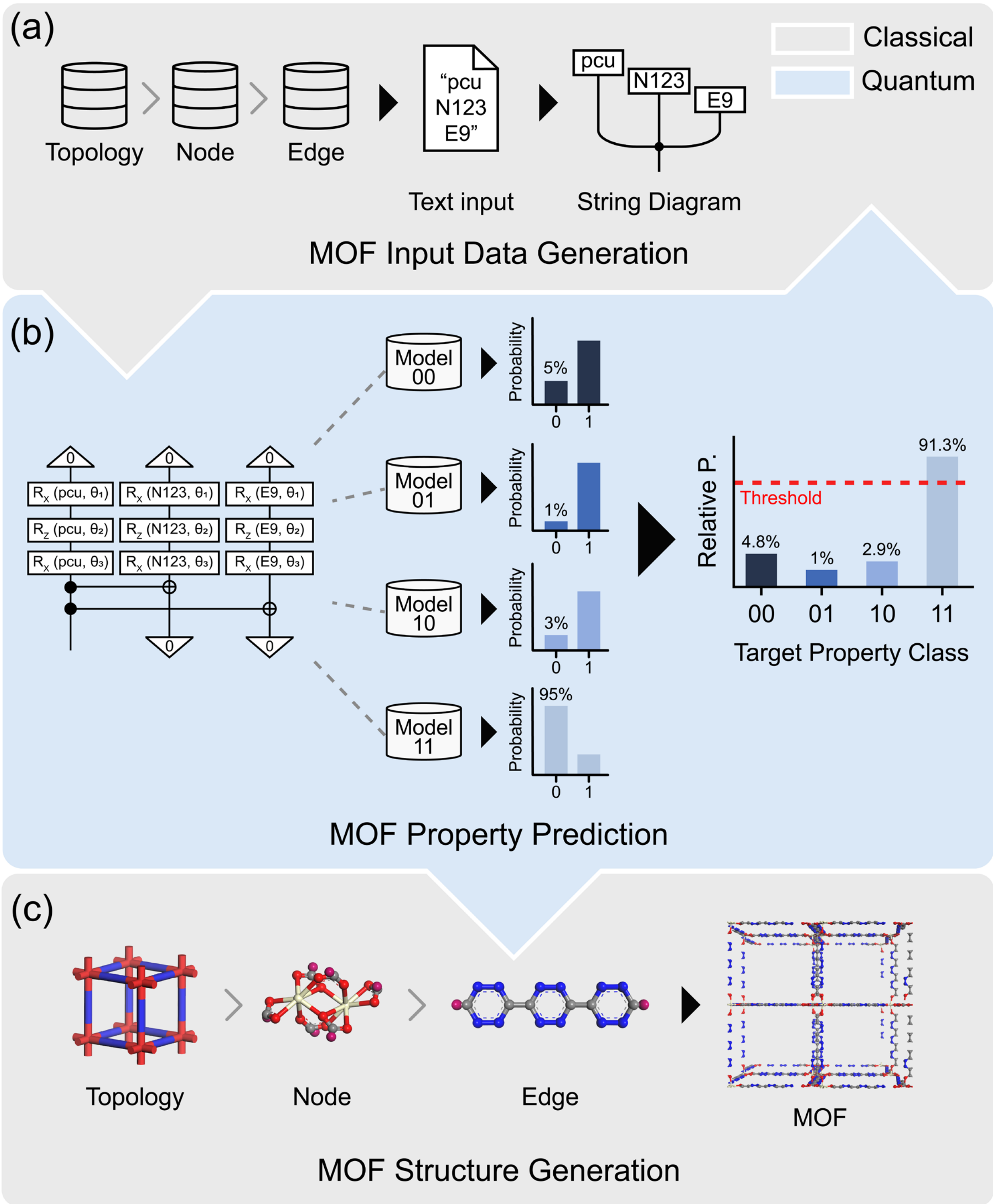

**Figure 7.** The overall workflow of the MOF generation process: **(a)** MOF input generation process based on random selection rule. **(b)** Property prediction based on relative probability distributions collected from four binary classification models. **(c)** MOF structure generation based on selected topology and building blocks.

The performance of the inverse design framework was evaluated based on 100 independent generation tasks for a given property class. To evaluate the inverse design performance more rigorously, the MOF search space for candidate generation was explicitly restricted to the test dataset only to ensure that none of the generated structures were seen during training. In this setup, the system was forced to generate candidate MOFs exclusively from previously unseen combinations. The resulting MOF generation framework showed good performance, with a total average accuracy of 93.88% (Table 1). The overall generation performance improved by 7.88% compared to the classification performance (Fig. 6). For both target properties, pore volume and $CO_2$ Henry's constant, the inverse design framework significantly improved the performance of the least accurate classification models, while models that already performed well in classification resulted in relatively small improvement. For example, the moderately high $CO_2$ Henry's constant class showed improvement of 20% compared to just 5% for the low $CO_2$ Henry's constant class. This improvement can be attributed to the collaborative multi-model architecture of the framework, where the evaluation of each MOF candidate incorporates contributions from all specialized QNLP classifiers. That is, the relative probability of the least performing classifier benefits from the contributions of other QNLP classifiers while the best-performing classifier risks its accuracy by incorporating predictions from relatively less accurate model. Consequently, the proposed framework maintained outstanding and consistent performance across target properties, achieving average accuracies of 97.75% for pore volume and 90% for $CO_2$ Henry's constant, even under the restricted test setup where candidate MOFs were generated exclusively from unseen combinations.

The average number of guesses until correct prediction can be interpreted as a metric that distinguishes between incorrect predictions and timeouts. A timeout occurs when the QNLP models' predictions reach 100 trials without achieving a prediction that satisfies the threshold probability. As the number of timeout trials increases, so does the average number of guesses, conveying a meaning slightly different from that of an incorrect prediction. An incorrect guess indicates the model's failure to accurately predict a MOF's property, while a timeout suggests that the model has not yet delivered its prediction, retaining the potential to eventually make correct predictions. This distinction allows for a comprehensive evaluation of the generation performance across target classes beyond merely assessing accuracy. Specifically, in the $CO_2$ Henry's constant dataset, the 'moderately low' class achieved 81% accuracy with an average of 28.44 guesses, whereas the 'high' class reached 90% accuracy with an average of 7.33 guesses (Table 1). When solely evaluating performance based on accuracy, the generation performance of the 'moderately low' case might appear inferior to that of the 'moderately high' case. However, the 'moderately low' case included 16 timeouts, while the 'high' case had none, meaning all 10 trials not included in the accuracy calculation were incorrect predictions (Table S4). As a result, our MOF generation framework of the multi-model-based architecture showed outstanding performance across target properties with an average accuracy of 97.75% and 90% for pore volume and $CO_2$ Henry's constant, respectively. Representative MOF structures generated by the inverse design process are illustrated in Fig. S12. The full set of 450 generated MOF structures has been provided in Crystallographic Information Framework (.cif) format and made publicly available via our GitHub repository, https://github.com/shinyoung3/MOF_QNLP.

**Table 1.** The test result of generating MOF with target property. The accuracy was calculated as the ratio of correct predictions over a total of 100 generation tasks (the number of correct, incorrect, and timeouts was considered).

| | **Pore Volume** | | **$CO_2$ Henry's Constant** | |
|---|---|---|---|---|
| **Target Property** | **Accuracy [%]** | **Avg. Guess** | **Accuracy [%]** | **Avg. Guess** |
| Low | 100 | 7.71 | 87 | 6.67 |
| Mod. Low | 100 | 10.73 | 81 | 28.44 |
| Mod. High | 91 | 9.98 | 98 | 25.13 |
| High | 100 | 11.76 | 90 | 7.33 |

## Extensibility of QNLP to Complex Reticular Frameworks

Although our simple MOF dataset of fixed set of topologies with a single building block combination resulted in a non-sequence and non-syntax-sensitive model as the optimum approach (i.e. BoW model), the authors want to highlight that this result doesn't capture the complexity inherent in MOF structures. For example, two mixed-metal MOFs with identical chemical compositions but different arrangements of the guest metal clusters can be identified as different structures because the spatial arrangement of the components can affect their chemical properties[41]. In fact, the sequence is one of the key parameters often introduced to compositionally repeating structures for structural representation and analysis. That is, the sequence of the materials becomes the boundary of determining whether they are identical or not. For example, Phong et al. addressed the periodicity of the block copolymers as a factor causing a difference in mechanical and thermal phase behavior[42]. They synthesized two types of dynamic block copolymers (DBCPs) with identical chemical compositions of PEG and PDMS oligomers but different sequences. Despite their identical chemical composition, well-ordered periodic DBCPs exhibited higher ionic conductivity and flow temperature than DBCPs with random sequences, which were attributed to the enhanced ion-transporting and thermal stability due to the formation of supramolecular nanofibers by stacking the hydrogen bonding units[42]. Another familiar example is DNA sequence. The sequence of DNA block copolymers was proved to play a crucial role in dictating the physical form and properties of the assembled structures as variations in the DNA sequences led to the formation of three distinctly different structures by Rizzuto et al[43].

As mentioned earlier, this concept is applicable to many other periodic crystalline materials, such as MOFs and covalent organic frameworks (COFs). Recently, Canossa et al. claimed the concept of sequence as one of the crucial indicators for defining structural identity and distinctive properties of multivariate crystals[44]. They defined this sequential property as unit cell information capacity (UCiC) to store tunable chemical or physical information of multivariate frameworks[44]. When a material has the number of $n$ tunable variables and $m_i$ is the multiplicity of $i^{th}$ variable, UCiC is defined as:

$$\mathrm{UCiC} = \prod_{i=1}^{n} m_i = \mathrm{m}_1 \times m_2 \times \ldots \times m_n \qquad (7)$$

The practical advantage of QNLP become particularly relevant when considering such complex multivariate frameworks based on its native tensor product operations and inherently probabilistic outputs. In our current study, each MOF is represented by a single topology, node, and edge combination, requiring only three qubits per structure. However, as structural complexity increases to accommodate multiple $n$ node and $m$ edge sites simultaneously (e.g. $topology - \mathrm{node}_1 - node_2 - \cdots - node_n - edge_1 - \mathrm{edge}_2 - \cdots - edge_m$), the distinction between QNLP and classical approaches becomes more pronounced. Classical methods would require explicit multiplication of weight matrices for each component by $MOF\ representation = w_{topology} \times w_{node_1} \times \ldots \times w_{node_n} \times w_{edge_1} \ldots \times w_{edge_m}$, whereas QNLP naturally encodes these relationships through quantum entanglement (e.g. CNOT gates) within a single state-preparation framework. The tensor product structure emerges naturally from quantum superposition by $|\psi_{MOF}\rangle = U_{gate}(|topology\rangle \otimes |node_1\rangle \otimes \ldots \otimes |node_n\rangle \otimes |edge_1\rangle \otimes \ldots \otimes |edge_m\rangle)$, where $|\psi_{MOF}\rangle$ is the quantum state that represents a single MOF after encoding, and $U_{gate}$ is the quantum circuit (a sequence of gates, including entangling gates) that prepares this state from the basis inputs ($|topology\rangle$, $|node_n\rangle$, $|edge_m\rangle$). This allows the joint composition is carried out by state preparation and entangling operations rather than by an explicit chain of classical matrix multiplications. Additionally, quantum measurements inherently produce normalized probability distributions, eliminating the normalization steps required in classical models (e.g. softmax).

The representational advantages of QNLP become increasingly relevant not only as structural complexity grows but also as representation richness increases. Our current simplified tokenization scheme (topology/node/edge + number), while convenient for proof-of-concept demonstration, is functionally close to one-hot encodings and does not capture deeper chemical meaning. The UCiC concept highlights how the multiplicity and chemical meaning of variables significantly impact material representation quality. Chemically meaningful tokens such as functional group descriptors (e.g. carboxylate, imidazole, sulfonate) or topology-based chemical encodings (e.g. octahedral_node, trigonal_planar) would encode richer semantic relationships that QNLP could exploit through its natural language processing capabilities. Such representations could enable QNLP to capture chemical intuition and learn similarities between chemically related components rather than treating each as completely independent entities. However, a trade-off exists between semantic richness and quantum resource requirements, as more complex tokenization schemes may demand deeper circuits or more sophisticated encoding strategies.

Our current focus on classification task provides a natural starting point for adapting QNLP circuits to material's data within NISQ hardware, while demonstrating the methodological feasibility. However, for practical inverse design that require quantitative property predictions such as precise adsorption energies or electrical conductivities, regression capabilities remain essential. A practical path forward is a two-stage framework. First, the QNLP classifier could rapidly screen promising candidates within desired property ranges, followed by a regression step using classical (or quantum) models or physics-based simulations. Indeed, recent quantum machine learning studies have begun adopting such hybrid workflows to navigate materials large design spaces by utilizing the representational advantages of quantum algorithms[45, 46].

As UCiC is a highly flexible variable contingent upon the framework's defined variables and variants, the optimal model for material representation also greatly depends on the complexity of the periodic structures. Consequently, the four compositional models we studied for representing simple MOF structures necessitate careful selection and further refinement based on the target materials. Thus, the extensibility of QNLP to other complex materials would depend on how well the algorithm respects the structural identity of the materials while addressing the statistical challenges from the increased requirement for qubits to process such complexities. While current NISQ hardware constrains demonstrations to relatively simple representations, the QNLP framework provides a foundation for extending to more complex multi-component systems as quantum resource scale.

## Discussion

In this study, we experimented with the applicability of quantum natural language processing for the inverse design of the MOFs with the desired target properties of pore volume and $CO_2$ Henry's constant. The target properties of pore volume and $CO_2$ Henry's constant were calculated against 450 hypothetical MOF structures consisting of 3 topologies, 10 metal nodes, and 15 organic edges. To construct a multi-class classification model, these datasets were evenly separated into four classes: low, moderately low, moderately high, and high classes for pore volume and $CO_2$ Henry's constant, respectively. A comparative study of pre-existing QNLP models was conducted to find an optimum approach to process quantum circuits in our MOF dataset. The results showed the best performance in the following order: BoW, DisCoCat, and sequence-based models, with the top-performing model achieving validation accuracies of 88.6% and 78% for binary classification based on the pore volume and $CO_2$ Henry's constant datasets, respectively. This performance is attributed to the non-sequence-dependent nature of our MOF datasets, meaning that completing a MOF structure relies on whether it includes a complete set of topology and building blocks rather than the order in which they are assembled. Based on the choice of the QNLP model, multi-class classification models were developed, considering the probabilistic nature of quantum circuit measurements. Four binary classification models were individually optimized to the specific target classes, and the models showed test accuracies of 92%, 92%, 86%, and 98% for low, moderately low, moderately high and high pore volume classes, respectively. Similarly, the test accuracies on the $CO_2$ Henry's constant dataset recorded 86%, 72%, 78%, and 84%. The performance of generating MOFs with desired properties showed improved accuracies by 7.88% on average owing to the requirements in the generation framework that the classification models exceed the confidence threshold.

Importantly, the goal of this study is not to demonstrate quantum computational advantage over classical machine learning methods in terms of speed, scalability, or accuracy. Instead, our aim is to establish a conceptual foundation for applying quantum computing, particularly in the current NISQ era, to modeling large molecular systems that go beyond the relatively simple periodic materials specifically explored in previous quantum machine learning studies. As discussed in the introduction, although NISQ devices offer potential for hybrid quantum-classical algorithms, clear demonstration of quantum advantage remain elusive due to qubit noise, small system sizes, and shallow circuit constraints[5, 10]. By exploring how QNLP encodes structural materials data such as MOFs, this study proposes a framework that can support future quantum-native strategies for materials design, particularly as quantum hardware continues to advance. While recent advance in de novo generative models for MOF inverse design, including diffusion[47], variational autoencoders (VAEs)[48], and transformers[49] models, have demonstrated remarkable success in proposing new MOF structures and chemistries, our approach focuses on property-guided selection within well-defined component spaces using quantum computational principles. The strategy is necessitated by fundamental scalability constraints, as MOF unit cells containing hundred of atoms would require prohibitively large number of qubits for direct quantum generative modeling. Consequently, our QNLP classifier-based framework serves as an exploration of quantum computing in materials informatics rather than competing directly with established generative models. Despite these limitations, applying QNLP to MOF design with limited datasets represents a meaningful first step toward quantum algorithm-based inverse design frameworks.

The extensibility of QNLP to other materials was further discussed based on the significance of sequence in effectively describing the complex nature inherent in the periodic materials. The comparative study presented in this paper was limited to the simple MOF structures, where the arrangement and sequence of components were considered less critical. However, this should not imply that all MOF data should be treated without regard to sequence. In fact, the complexity of many periodic materials necessitates careful design that considers sequence as a key factor in defining structural identity[44]. This leads us to believe that the effectiveness of QNLP in material science depends on its ability to deal with the detailed structures of the materials. Although our study only looked at a small part of the wide variety of MOF search spaces, we believe applying QNLP to design MOFs with a limited dataset is an exciting first step. Our approach, inspired by successful applications of QNLP in fields as diverse as sentence generation[15] and music composition[16], demonstrated the potential for classifying and generating large periodic materials with desired properties. This method could provide a novel perspective on efficiently navigating the vast search space of MOFs by bridging the gap between quantum algorithm and material design.

# Methods

## Dataset Preparation

The MOF structures for the quantum model training were generated using our in-house software, PORMAKE[40]. These 450 hypothetical MOF structures were optimized by applying a force field in BIOBIA Materials Studio 2019[50]. The target geometrical property, pore volume, was calculated based on a channel and probe radius of 1.2 and the number of Monte Carlo samples per unit cell of 5000 using Zeo++[51]. The target chemical property, $CO_2$ Henry's constant, of 450 MOFs was calculated using Monte Carlo simulations in the RASPA package[52]. The simulations were performed at 298K using the Widom insertion method, which estimates Henry's constant by inserting test molecules into the MOF structures at infinite dilution. UFF was used for MOF structures, applying Ewald summation for electrostatics with an Ewald precision of $10^{-6}$ and a cutoff distance of 12.8 Å. The Lorentz-Berthelot mixing rule was used to determine the interaction parameters between framework atoms and $CO_2$ molecules. Each simulation ran for 50,000 Monte Carlo cycles, with Henry's constant computed from the Widom insertion results. Then, boundaries for class labels were determined based on 25, 50, and 75 percentiles of target properties' distributions (Fig.2). Train, validation, and test datasets were divided into 300, 100, and 50 samples.

## Quantum Model Training

The MOF input data was processed for quantum computation using lambeq, an open-source Python library for QNLP[37]. The input data consisted of a class label and its corresponding MOF name in order of topology, node, and edge (Supplementary Information 2). The difference between NLP and QNLP starts from the input data processing, while their goal remains the same: translation between unstructured and structured language. To prepare the quantum circuits representing instructions of the gate operations to encode MOF information in qubits, the dataset was first tokenized and processed into the string diagrams based on the choice of the models: bag-of-word, DisCoCat, and word-sequence models. The four types of method were used to generate the string diagrams as shown in Fig. 2 and Fig. S3. The DisCoCat model-based approach necessitates the conversion of pregroup diagrams into string diagrams; accordingly, the dataset underwent processing based on rewriting rules[35]. The string diagrams, representing MOF data, were subsequently transformed into quantum circuits employing the IQP (Instantaneous Quantum Polynomial) ansatz with the choice of two Rx gates and a single Rz gate. Although the specific selection of the ansatz should not be overly emphasized[35], the IQP was chosen due to its well-reported accomplishments in prior research[15, 16, 35, 53]. The single IQP layer was used throughout the experiments to maintain the circuit depth small.

Importantly, the circuit architecture differs between binary and multi-class classification tasks due to how the lambeq software internally handles single- and multi-qubit representations. For single qubit representation (as used in binary classification), lambeq allows multiple parameterized gates such as Rx-Rz-Rx arrangement per qubit, controlled by the n_single_qubit_params argument. This explicit multi-parameterization helps avoid degeneracy issues common in single-qubit circuits. However, when components are represented using multiple qubits (as in multi-class classification), lambeq internally applies a fixed entangled architecture with Hadamard and controlled gates, and systematically disables multiple single-qubit rotations (assigning only one trainable parameter per qubit). Degeneracy in this case is handled naturally via entablement. The number of parameters for multi-qubit representations, therefore, can be only controlled by replicating the circuit depth (Fig. S7). Since we adopted the single circuit layer approach throughout this study, this design decision affects the total number of circuit parameters and is further discussed in more detail in Note S4.

All QNLP models studied in this paper were trained using simulated hardware based on Qiskit AerBackend[26] with 8192 shots (2048 shots for binary models 00 to 11). In addition, all models were trained on a cross-entropy objective, and the quantum parameters were updated based on the Simultaneous Perturbation Stochastic Approximation (SPSA) optimization method[54]. The initial learning rate and initial parameter shift scaling factor were set as 0.05 and 0.06, respectively, while the stability constant of the classical optimizer were uniquely chosen for each model based on the comparative analysis test (Table S5).

## MOF Text Input Generation Algorithm

The generation algorithm was implemented with context-free grammar (CFG) using the natural language toolkit[55]. The CFG for MOFs was designed to adhere to a random generation rule. Individual MOF building blocks were randomly selected from the search space for topology, nodes, and edges, respectively.

## MOF Generation Test

The MOF generation performance was measured based on 100 individual tests for each class. The accuracy of generating MOF with the target class was calculated based on the ratio of correct guesses against total guesses. Total guesses were defined as a sum of time-out, correct, and incorrect guesses (i.e. 100 runs). The time-out indicates a situation in which the QNLP model's prediction exceeds 100 iterations to output a prediction satisfying a threshold probability of 85% for the pore volume dataset and 65% for the $CO_2$ Henry's constant dataset.

## Data availability

Text datasets used for model training and testing, as well as MOF structure files (.cif), are available from https://github.com/shinyoung3/MOF_QNLP.

## Code availability

Code for MOF QNLP model training and MOF generation is available from https://github.com/shinyoung3/MOF_QNLP

## Acknowledgements

We thank National Research Foundation of Korea (Project Number RS-2024-00337004) for the financial support.

## Author Contributions

S.K. performed the calculation, analysis and writing of the manuscript and J.K. supervised the project. All authors read and approved the manuscript.

## Competing Interest

The authors declare no competing financial interest.

# Figure Legends

**Figure 1.** General overview of the study. Gray, blue, and yellow boxes indicate the classical computer-based, quantum computer-based, and hybrid quantum-classical computer-based processes, respectively. **(a)** MOF dataset construction from target property distributions encoded into four classes, and it is transformed into the text dataset consisting of the target class and its corresponding MOF name. **(b)** Comparative study among QNLP model candidates. Models A, B, C, and D refer to DisCoCat, bag-of-words (BoW), word-sequence, and stair models, respectively. **(c)** Multi-classification model development based on the choice of QNLP model from the QNLP model selection step. Four binary classification models by target property type (pore volume and $CO_2$ Henry's constant) are combined to complete a multi-classification framework. **(d)** MOF generation framework to build MOF structure with the desired target property.

**Figure 2.** Distributions of 450 hypothetical MOFs based on **(a)** pore volumes and **(c)** the logarithm (base 10) of $CO_2$ Henry's constants (The unit of $CO_2$ Henry's constants is mol/kg/Pa). The datasets were divided into four classes based on 25th, 50th, and 75th percentiles of the pore volume and $CO_2$ Henry's constant values. Vertical

dotted lines indicate corresponding class boundaries. Counts of classes per topology type, node (metal clusters) type, and edge (organic ligands) type based on **(b)** pore volumes and **(d)** $CO_2$ Henry's constants. Labels 00, 01, 10, and 11 represent low, moderately low, moderately high, and high pore volumes and $CO_2$ Henry's constants, respectively. Vertical red dotted lines represent a count boundary of the class-significance.

**Figure 3. (a)** String diagrams of MOF representation based on BoW model, and **(b)** its corresponding quantum circuit with a single-qubit configuration for binary classification based on IQP ansatz. Dotted line boxes represent **(i)** initial quantum states, **(ii)** circuit parameters and **(iii)** controlled-not gates which indicating merge-dot. Open-wire that carries predicted class label is highlighted as blue line. **(c)** An overall process of quantum machine learning.

**Figure 4.** Training dynamics and final accuracies and losses for four QNLP models (the bag-of-words (BoW), DisCoCat, word-sequence with Cups (Cups), and word-sequence model with stairs (Stairs) models) trained on **(a)** pore volume and **(b)** $CO_2$ Henry's constant dataset for binary classification task. The left-hand panels display the training losses and accuracies as a function of epochs, showing convergence behavior. The right-hand panels summarize the final training and validation looses (top) and accuracies (bottom) for each model.

**Figure 5. (a)** Quantum circuits of MOF named pcu N248 E70 with two-qubit configuration for multi-classification tasks. Note that the blue lines represent open wires, and triangles pointing downward represent qubits with a 0-effect. **(b)** A post-selection process assuming the quantum circuit measurement result in a uniform distribution for binary classification and **(c)** multi-classification tasks. The circuit measurement outcomes after the post-selection is vectorized to be matched with one-hot encoding form of the true labels.

**Figure 6.** Training history of loss based on **(a)** pore volume and **(b)** $CO_2$ Henry's constant dataset. Accuracy bar graphs of **(c)** pore volume and **(d)** $CO_2$ Henry's constant datasets. Note that Multi, $Bi_{00}$, $Bi_{01}$, $Bi_{10}$, and $Bi_{11}$ represent BoW models trained based on the multi-classification dataset, 00-specialized dataset (binary classification dataset divided into classes 00 and 01+10+11), 01, 10, and 11-specialized datasets, respectively.

**Figure 7.** The overall workflow of the MOF generation process: **(a)** MOF input generation process based on random selection rule. **(b)** Property prediction based on relative probability distributions collected from four binary classification models. **(c)** MOF structure generation based on selected topology and building blocks.

## Author information

### Corresponding Author

*Jihan Kim – Department of Chemical and Biomolecular Engineering, Korea Advanced Institute of Science and Technology (KAIST), E-mail: jihankim@kaist.ac.kr

### Present Addresses

Shinyoung Kang: Korea Advanced Institute of Science and Technology, Daejeon, 34141, Republic of Korea

Jihan Kim: Korea Advanced Institute of Science and Technology, Daejeon, 34141, Republic of Korea

# *Supporting Information:*

## Property-Guided Inverse Design of Metal-Organic Frameworks Using Quantum Natural Language Processing

*Shinyoung Kang, Jihan Kim**

Department of Chemical and Biomolecular Engineering, Korea aAdvanced Institute of Science and Technology
291 Daehak-ro, Yuseong-gu, Daejeon 34141, Republic of Korea

# Table of Contents

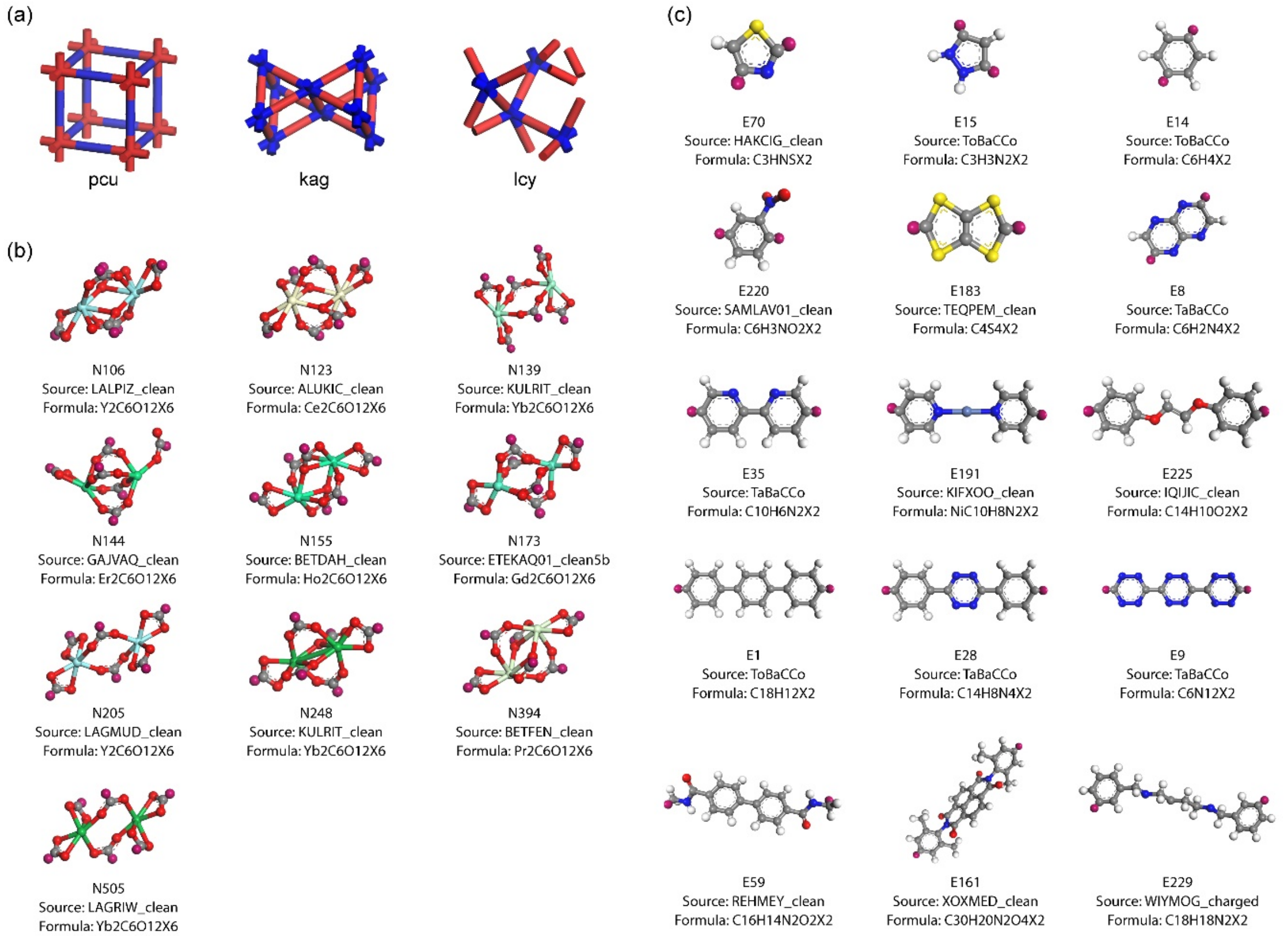

**Figure S1.** Topologies and building blocks of MOF dataset: **(a)** topology, **(b)** node, and **(c)** edge. X indicates the connection point with building blocks.

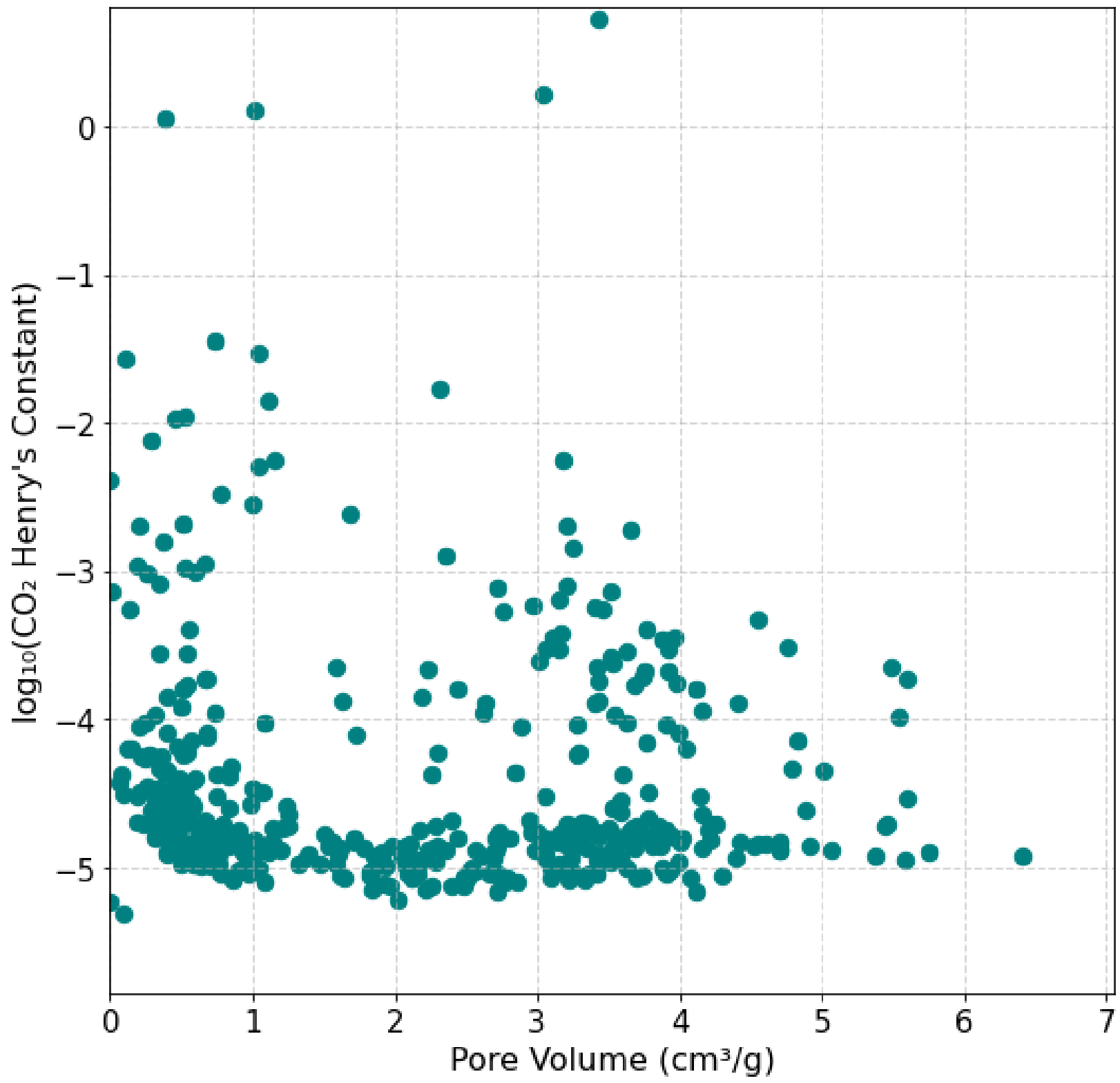

**Figure S2.** Scatter plot showing a correlation between pore volume and $CO_2$ Henry's constant (The unit of $CO_2$ Henry's constants is mol/kg/Pa).

**Note S1: MOF Representation based on QNLP Models**

Given that our MOF dataset is organized by its modular components (i.e. topology, node, and edge) similarly to how sentences are formed from a selection of words, it is well-suited for the direct application of various QNLP models to construct string diagrams, the quantum version of tensor networks. The DisCoCat model processes the data based on Combinatory Categorial Grammar (CCG)[1], which is designed for linguistic parsing, and it focuses on the syntactic and semantic relationships between words in a sentence. In this context, a MOF can be viewed as a noun phrase atomic type, $n$, where the edge acts as a relative pronoun atomic type, $n$, and the node and the topology function as a left adjoint (i.e., $n.l$) and a right adjoint (i.e., $n.r$), respectively (Fig. S3a). The pregroup grammar in DisCoCat has two reduction rules[2]:

$$\mathrm{p} \cdot \mathrm{p}^r \rightarrow 1 \qquad \mathrm{p}^l \cdot \mathrm{p} \rightarrow 1 \qquad (1)$$

where p denotes atomic type. Thus, the pregroup string diagram for MOF such as "pcu N248 E70" is represented by $(n \cdot (n^l \cdot n) \cdot n^r) \cdot n$ and the derivation for single MOF is reduced to $(n \cdot (n^l \cdot n) \cdot n^r) \cdot n \rightarrow n \cdot 1 \cdot n^r \cdot n \rightarrow (n \cdot n^r) \cdot n \rightarrow 1 \cdot n \rightarrow n$ based on aforementioned reduction rules. Although its pregroup grammar was not designed for chemical structures like MOFs, we decided to include this fully-syntax sensitive model for comparative purposes. The transition from these pregroups to vector space semantics is processed by mapping each type to a vector space ($n$ to a vector space $N$) and composite types to corresponding tensor product spaces ($n^r \cdot n$ to a tensor product space $N \otimes N$)[3]. As shown in Fig. S3a, boxes in the DisCoCat string diagram represent tensors, the order of which is determined by the number of their wires, and the cups (U-shape wire denoting $n^l \cdot n$ or $n \cdot n^r$) are the tensor contraction governed by the reduction rules. After applying reduction rules, the string diagram is further simplified as shown in Fig. S4. The transformation to the quantum circuit is achieved by the selection of ansatz and each MOF component vector being mapped into individual qubits (Fig. S3d). The specific arrangement of unitary operations in the circuit will not be discussed since this is beyond the scope of this paper, but it is basically based on the compositional and reduction rules from the DisCoCat model.

Lastly, two types of word-sequence approaches were considered, and these models are the quantum version of a simple recurrent neural network (RNN). The word-sequence model captures the order of the MOF components. A tensor, $s$, represents each MOF component with the multiplication of the corresponding adjoint type, s.r, and the sequential multiplication of these tensors in the order becomes the MOF as the output vector. This process begins with the "start" token vector and the connectivity of each component with left-to-right sequence is represented as cups as demonstrated in Fig. S3b. Here, all component vectors, including the "start" vector, are mapped into the individual qubits, and the unitary operations reflect the MOF structure as defined by the sequential multiplication of its components (Fig. S3e). Another word-sequence approach available in lambeq[2] involves the use of a “stair” instance, which executes the tensor multiplication sequentially from left to right (Fig. S3c). In this case, the stair instance is encoded as a unitary operation within the quantum circuit. So this approach can be an economical option since it generates the output without the additional “start” token and the adjoint vectors (Fig. S3f).

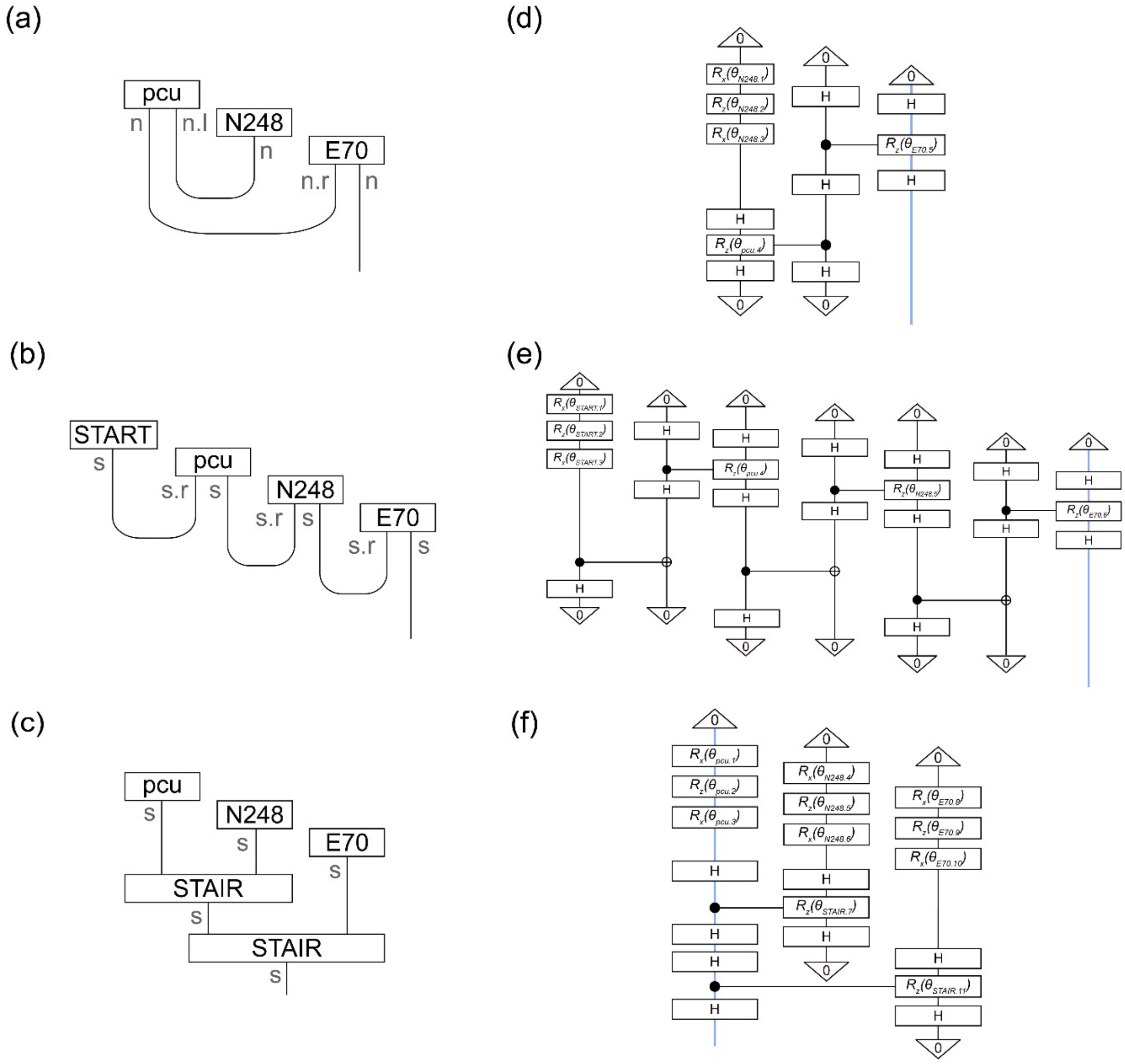

**Figure S3.** String diagrams of MOF representation based on **(a)** DisCoCat model, **(b)** word-sequence model, and **(c)** word-sequence model with a stairs-like recurrent manner. Their corresponding quantum circuits based on IQP ansatz were shown in **(d)**, **(e),** and **(f)** for DisCoCat, word-sequence, and stair models, respectively. Open-wire that carries predicted class label is highlighted as blue line.

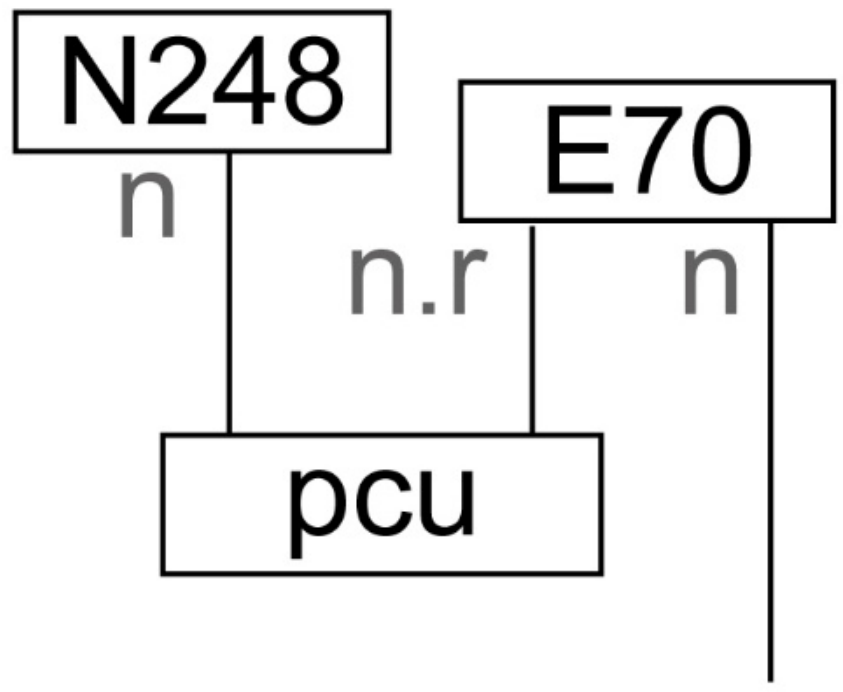

**Figure S4.** String diagram of MOF ‘pcu N248 E70’ based on DisCoCat model after applying reduction rules.

**Note S2: Identification of Optimal MOF Dataset Sequences for QNLP Models**

To determine the optimal input sequence for the sequence-sensitive QNLP models, we considered all possible permutations of the MOF components (topology (T), node (N), and edge (E)). This resulted in six unique sequences, ENT, ETN, NET, NTE, TEN, and TNE. Figure S5 illustrates how variations in the input sequence affected the performance of the Bag-of-Words (BoW) model, DisCoCat model, the word-sequence model with Cups (Cups model), and the word-sequence model with stairs (Stairs model) to input sequence variations across both the pore volume and $CO_2$ Henry's constant datasets. For clarity, only training accuracies are presented in Figure S5, as the validation accuracies followed similar trends. Complete numerical results for both training and validation datasets are provided in Table S1.

For the DisCoCat model, the ETN sequence achieved the highest training and validation accuracies (91.2% and 91%, respectively) with the lowest associated loss in the pore volume dataset. For the $CO_2$ Henry's constant dataset, the ENT sequence yielded the best performance, achieving 70% training accuracy and 69% validation accuracy, and the lowest loss. This indicates a notable sensitivity of the DisCoCat model to input sequence, possibly due to its syntax-aware structure, which reflects a dependence on input ordering. The stairs model exhibited a distinct behavior wherein datasets sharing the same initial component displayed identical performance metrics. For instance, ENT and ETN sequences produced the same accuracies and loss values. For the pore volume dataset, sequences starting with nodes (NET and NTE) showed superior performance. Conversely, sequences starting with topology (TEN and TNE) provided better results for the $CO_2$ Henry's constant dataset. This trend shows that the initial component significantly influences the performance of the Stairs model. The cups model consistently showed low performance among the sequence-based models, indicating that its specific tensor multiplication scheme, which relies on adjoint vectors and start tokens to represent the input sequence, might not efficiently capture MOF structural characteristics. Nonetheless, the optimal sequences identified for the Cups model were NTE for the pore volume dataset and TNE for the $CO_2$ Henry's constant dataset. As a result, the optimal input sequences were successfully identified for each QNLP model, enabling further comparative analysis of their performance.

The observed sensitivity of QNLP models to the sequence of MOF components arises primarily from the way quantum circuits encode and process input information. In sequence-sensitive architectures such as DisCoCat, Cups, and Stairs, input qubits are processed through layers of fixed quantum gate operations. Each MOF component, topology, node, and edge, is assigned to a specific qubit, and the order in which they appear determines how quantum gates are applied to encode their interactions. Consequently, the same components placed in different positions within the sequence undergo different unitary transformations, which alters the resulting quantum state, and in turn, the models' prediction. For example, in the DisCoCat model, the topology component "pcu" in the TNE sequence "pcu N248 E70" undergoes a sequence of three parameterized gates, Rx-Rz-Rx gates (Fig. S6a), whereas the same "pcu" component, when placed in the middle position in NTE sequence "N248 pcu E70", is only encoded by a single Rx gate (Fig. S6d). This difference in the number and type of applied gates leads to distinct quantum state representations, explaining why DisCoCat showed the largest variation in performance across different component sequences. In fact, DisCoCat exhibited a 26% gap in a mean accuracy between its best-performing (ETN) and worst-performing (NTE) sequences. Similar trends are observed in the Stairs model, where changes in the component sequences also lead to different entanglement patterns (compare Figs S6c and S6f). The Stairs model showed a 20% difference in mean accuracies between the most and least favorable sequences. In contrast, the Cups model, which uses a more uniform gate application structure across qubits regardless of order (Figs. S6b and S6e), exhibited only a 5% mean difference in performance, suggesting that it is relatively more robust to sequence permutations.

As expected, the BoW model demonstrated sequence independence for both datasets. For the pore volume datasets, all sequence permutations yielded identical training and validations accuracies at 0.925 and 0.93 with losses of 0.169 and 0.149, respectively. Similarly, for the $CO_2$ Henry's constant dataset, all sequences produced identical training and validations accuracies at 0.807 and 0.78 with losses of 0.443 and 0.523, respectively. These results align with our theoretical expectations for the BoW model, which relies on component-wise multiplication rather than sequential processing. Importantly, the BoW model's consistent performance across all sequences was superior to the best-performing configurations of sequence-dependent models. Even when the sequence-dependent models were optimized with their best sequence orders (e.g. ETN for DisCoCat, NTE for Cups, and NET/NTE for Stairs in pore volume), the BoW model achieved higher accuracies. This performance advantage was maintained regardless of sequence ordering, supports our conclusion about the suitability of component-wise representation for MOF design.

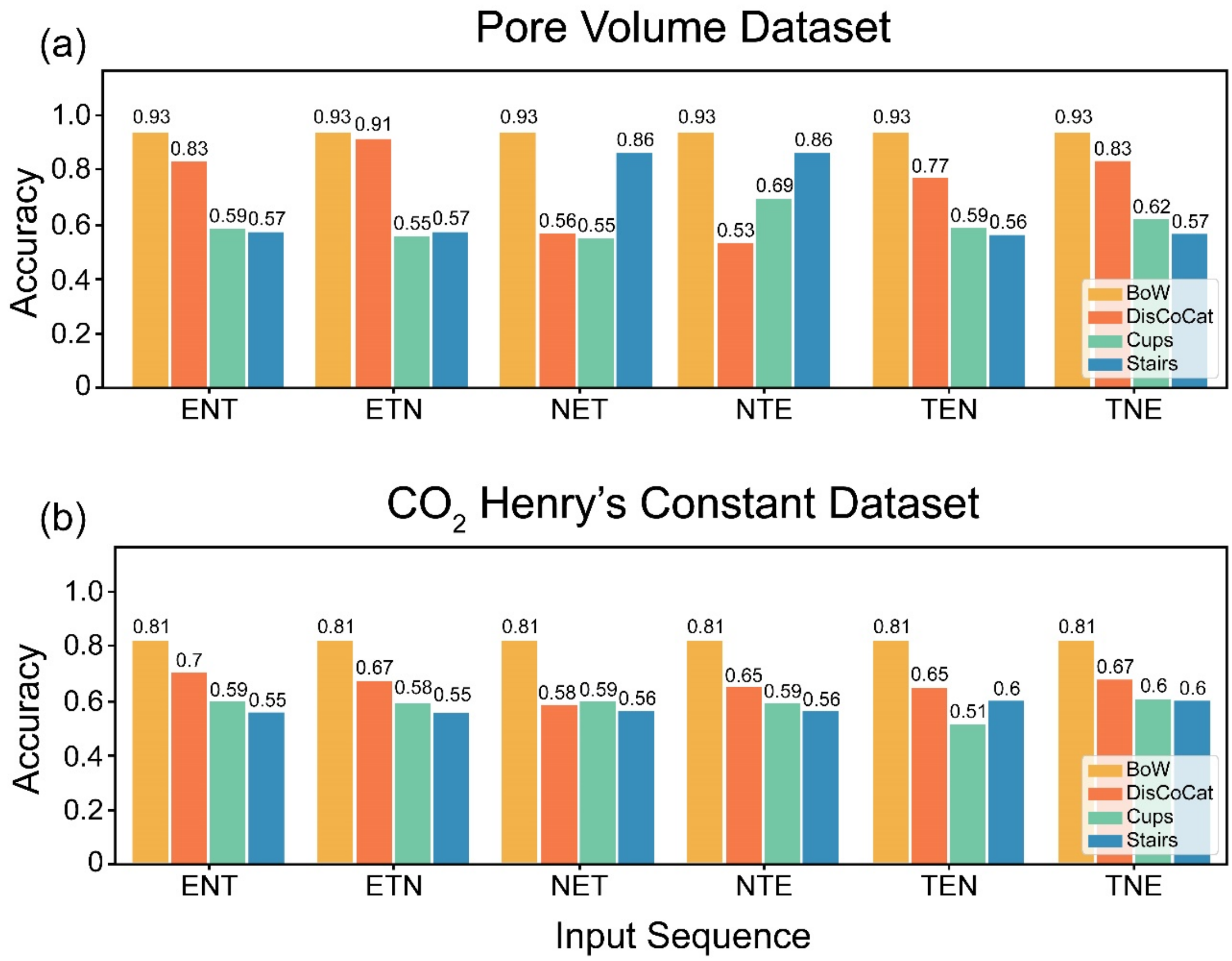

**Figure S5.** The effects of input sequence on the training accuracy of different QNLP models (Bag-of-Words (BoW), DisCoCat, sequence model with cups (Cups), and sequence model with stairs (Stairs)) for (a) pore volume dataset and (b) CO2 Henry's constant dataset. Six different input sequences were tested by permuting the order of MOF components, edge (E), node (N), and topology (T), resulting in the combinations of ENT, ETN, NET, NTE, TEN, and TNE.

**Table S1.** Training and validation accuracies and losses for four QNLP models (Bag-of-Words (BoW), DisCoCat, Sequense with Cups (Cups), Sequence with Stairs (Stairs)) evaluated on pore volume and CO2 Henry's constant datasets with varying input sequence orders. Six unique sequences of MOF components (Edge (E), Node (N), Topology (T)) were assessed to identify the optimal order that yields the highest performance.

| Dataset | Model Type | Input Sequence | Train Accuracy | Validation Accuracy | Train Loss | Validation Loss |
|---|---|---|---|---|---|---|
| Pore volume | BoW | ENT | 0.925 | 0.930 | 0.169 | 0.149 |
| | BoW | ETN | 0.925 | 0.930 | 0.169 | 0.149 |
| | BoW | NET | 0.925 | 0.930 | 0.169 | 0.149 |
| | BoW | NTE | 0.925 | 0.930 | 0.169 | 0.149 |
| | BoW | TEN | 0.925 | 0.930 | 0.169 | 0.149 |
| | BoW | TNE | 0.925 | 0.930 | 0.169 | 0.149 |
| | DisCoCat | ENT | 0.830 | 0.810 | 0.427 | 0.487 |
| | DisCoCat | ETN | 0.912 | 0.910 | 0.217 | 0.299 |
| | DisCoCat | NET | 0.563 | 0.555 | 0.684 | 0.694 |
| | DisCoCat | NTE | 0.528 | 0.510 | 0.683 | 0.709 |
| | DisCoCat | TEN | 0.770 | 0.660 | 0.520 | 0.675 |
| | DisCoCat | TNE | 0.833 | 0.850 | 0.435 | 0.388 |
| | Cups | ENT | 0.585 | 0.610 | 0.743 | 0.684 |
| | Cups | ETN | 0.553 | 0.615 | 0.687 | 0.680 |
| | Cups | NET | 0.547 | 0.625 | 0.751 | 0.687 |
| | Cups | NTE | 0.693 | 0.690 | 0.623 | 0.880 |
| | Cups | TEN | 0.585 | 0.590 | 0.718 | 0.670 |
| | Cups | TNE | 0.618 | 0.629 | 0.671 | 0.609 |
| | Stair | ENT | 0.570 | 0.530 | 0.673 | 0.718 |
| | Stair | ETN | 0.570 | 0.530 | 0.673 | 0.718 |
| | Stair | NET | 0.860 | 0.880 | 0.419 | 0.392 |
| | Stair | NTE | 0.860 | 0.880 | 0.419 | 0.392 |
| | Stair | TEN | 0.558 | 0.450 | 0.726 | 0.738 |
| | Stair | TNE | 0.565 | 0.486 | 0.694 | 0.706 |
| $CO_2$ Henry's Constant | BoW | ENT | 0.807 | 0.780 | 0.443 | 0.523 |
| | BoW | ETN | 0.807 | 0.780 | 0.443 | 0.523 |
| | BoW | NET | 0.807 | 0.780 | 0.443 | 0.523 |
| | BoW | NTE | 0.807 | 0.780 | 0.443 | 0.523 |
| | BoW | TEN | 0.807 | 0.780 | 0.443 | 0.523 |
| | BoW | TNE | 0.807 | 0.780 | 0.443 | 0.523 |
| | DisCoCat | ENT | 0.700 | 0.690 | 0.586 | 0.983 |
| | DisCoCat | ETN | 0.667 | 0.630 | 0.575 | 0.795 |

| | | | | | |
|---|---|---|---|---|---|
| DisCoCat | NET | 0.582 | 0.540 | 0.667 | 0.710 |
| DisCoCat | NTE | 0.647 | 0.685 | 0.643 | 0.631 |
| DisCoCat | TEN | 0.642 | 0.610 | 0.669 | 1.137 |
| DisCoCat | TNE | 0.673 | 0.615 | 0.604 | 0.909 |
| Cups | ENT | 0.593 | 0.515 | 0.678 | 0.705 |
| Cups | ETN | 0.580 | 0.555 | 0.673 | 0.678 |
| Cups | NET | 0.593 | 0.515 | 0.678 | 0.705 |
| Cups | NTE | 0.587 | 0.610 | 0.657 | 0.616 |
| Cups | TEN | 0.510 | 0.495 | 1.049 | 0.989 |
| Cups | TNE | 0.603 | 0.645 | 0.761 | 1.001 |
| Stair | ENT | 0.553 | 0.470 | 0.682 | 0.694 |
| Stair | ETN | 0.553 | 0.470 | 0.682 | 0.694 |
| Stair | NET | 0.560 | 0.440 | 0.675 | 0.715 |
| Stair | NTE | 0.560 | 0.440 | 0.675 | 0.715 |
| Stair | TEN | 0.597 | 0.485 | 0.687 | 0.713 |
| Stair | TNE | 0.597 | 0.485 | 0.687 | 0.713 |

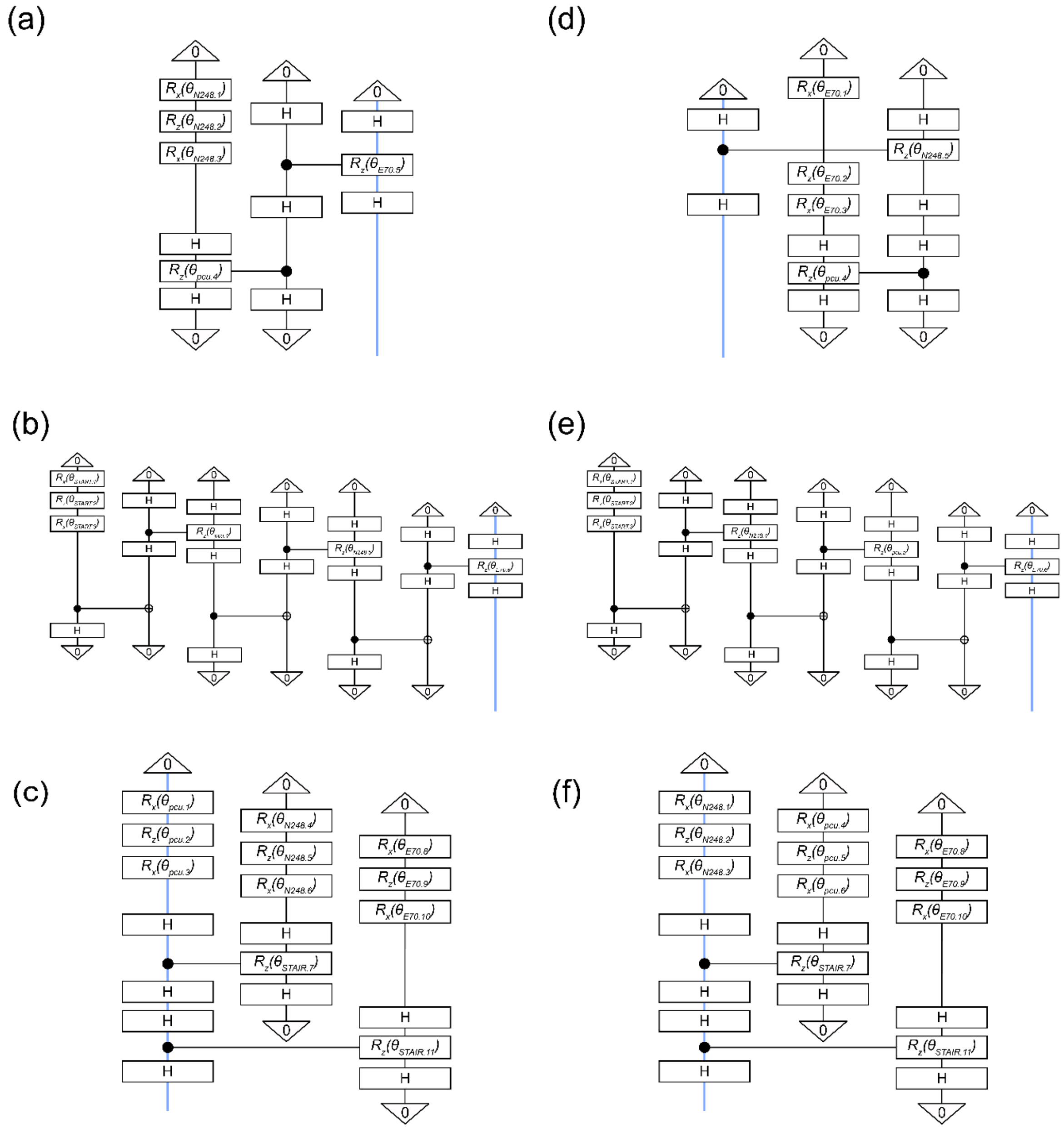

**Figure S6.** Quantum circuits for different MOF component sequences using DisCoCat, Cups, and Stair models. Quantum circuits for the TNE sequence (“pcu N248 E70”) using **(a)** DisCoCat, **(b)** Cups, **(c)** Stairs models, respectively. Quantum circuits for the NTE seuquence (“N248 pcu E70”) using **(d)** DisCoCat, **(e)** Cups, **(f)** Stairs models, respectively

**Note S3: Comparative Benchmarking of QNLP and Conventional Classical Machine Learning Models for MOF property Prediction**

In this study, our primary goal was to explore and validate the feasibility of applying QNLP techniques to MOF design tasks, particularly under the constraints of current NISQ hardware. As noted, the QNLP-based approach is not yet expected to surpass state-of-the-art classical models in raw predictive accuracy, particularly due to hardware limitations and the early stage of QNLP algorithmic development.

However, we recognize that a baseline comparison with conventional ML approaches can provide useful context. Therefore, we have conducted an additional benchmarking experiment using two widely adopted classical ML models, Random Forest classifier and Support Vector Machine (SVM) with the RBF kernel. For the comparison, we used the same MOF datasets for binary classification of pore volume and $CO_2$ Henry's constant, with identical train, validation, and test splits, and applied the same performance metrics as for the quantum ML (QML) model. MOF structures were converted into classical bag-of-words feature vectors using CountVectorizer from scikit-learn, and model hyperparameters were kept at their default settings. Since the quantum model does not support cross-validation, we instead varied the random_state parameter for the classical models over 10 independent runs to estimate mean and standard deviation of validation and test accuracies.

The result of this comparison, including validation/test accuracies and model complexity, is provided in Table S2. The classical models achieved higher validation accuracies, as expected, but the QNLP classifier performed competitively despite relying on significantly fewer trainable elements and a fundamentally different computational paradigm. Surprisingly, the QNLP classifier showed superior test accuracies for both pore volume and $CO_2$ Henry's constant properties, compared to conventional ML models, indicating strong generalization performance. The relatively higher standard deviation of QML model stems from the inherent probabilistic nature of quantum measurements. Unlike classical models that produce deterministic outputs from fixed weights and inputs, quantum models relay on the measurement of qubit states, which yield probabilistic outcomes governed by the underlying quantum circuit. This means that even when using the same trained circuit and input data, repeated executions (shots) of the quantum circuit can result in slight variations in output class probabilities. Despite this variability, the QNLP model consistently demonstrated competitive average accuracy across multiple runs, showing its potential as a viable modeling approach for MOF property prediction.

We also emphasize that a direct comparison of accuracy can be somewhat misleading because classical and quantum approaches differ in representational principles. As discussed in the main manuscript, QNLP encodes compositional structure natively into the circuit using entangled qubit states, whereas classical models rely on engineered feature or embeddings, requiring extensive vector multiplications and additional post-processing steps. The QNLP circuits, in contrast, produce probabilistic predictions directly from quantum measurements, offering a more compact and interpretable representation of structural correlations within MOFs. While we acknowledge that QNLP is not yet a replacement for mature classical ML pipelines, we believe it provides a compelling novel architecture, particularly as quantum hardware continues to advance.

**Table S2.** Comparison of classical machine learning models (Random Forest and Support Vector Machine with RBF kernel) and the quantum natural language processing (QNLP) classifier on MOF property classification tasks. Metrics include average validation and test accuracy, standard deviation over 10 independent runs, and model complexity in terms of number of decision trees, support vectors, or quantum circuit parameters.

| **Property** | **Dataset** | **Metric** | **Random Forest** | **SVM (RBF kernel)** | **QNLP classifier** |
|---|---|---|---|---|---|
| Pore volume | Validation | Accuracy [%] | 94.3 | 94 | 90.86 |
| | | Standard Dev. [%] | 0.5 | 0 | 1.14 |
| | Test | Accuracy [%] | 88 | 86 | 90.33 |
| | | Standard Dev. [%] | 0 | 0 | 3.79 |
| | Training | Model Complexity | Num. of trees: 100<br>Total nodes: 20,004 | Num. of support vector: 246 | Num. of circuit parameters: 84 |
| $CO_2$ Henry's constant | Validation | Accuracy [%] | 81.9 | 83 | 77 |
| | | Standard Dev. [%] | 0.92 | 0 | 1.1 |
| | Test | Accuracy [%] | 71.8 | 74 | 77.8 |
| | | Standard Dev. [%] | 1.66 | 0 | 1.08 |
| | Training | Model Complexity | Num. of trees: 100<br>Total nodes: 20,004 | Num. of support vector: 246 | Num. of circuit parameters: 84 |

(a)

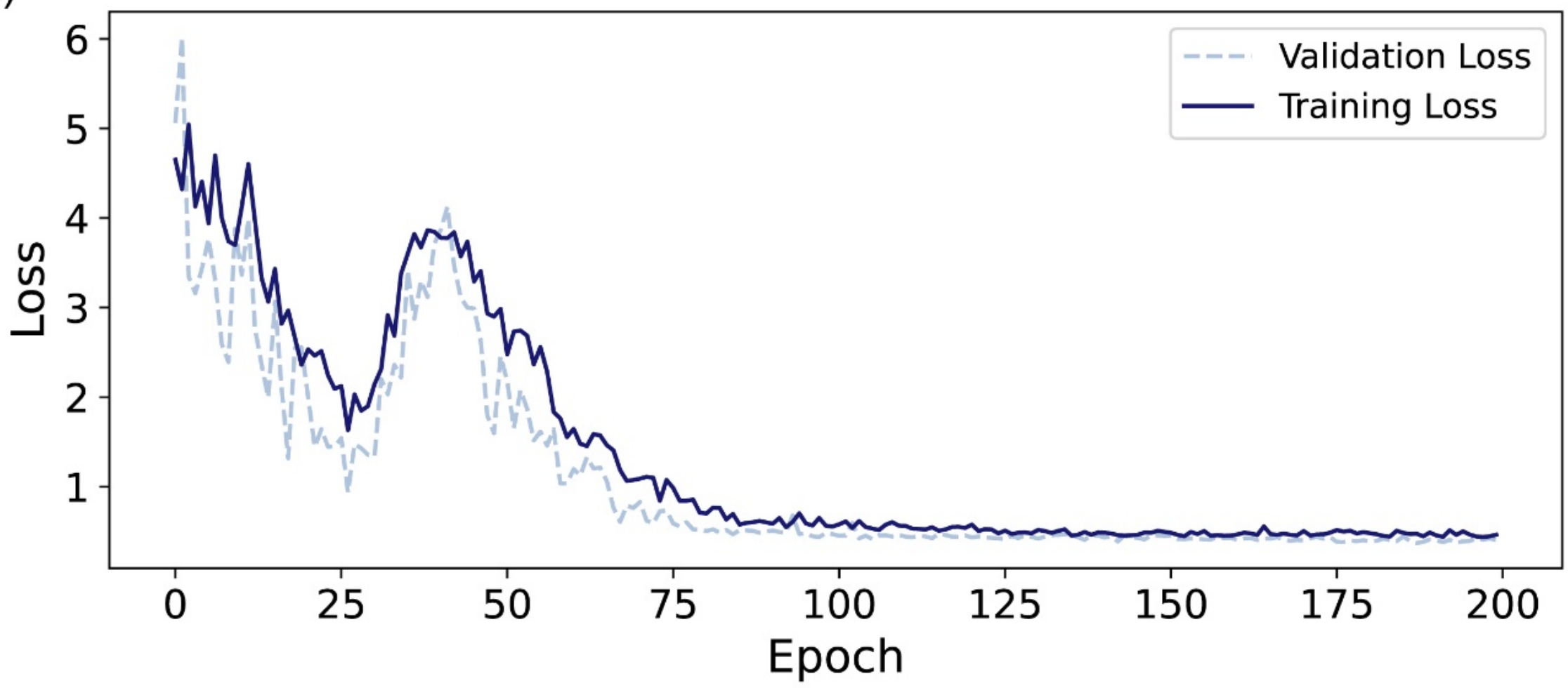

(b)

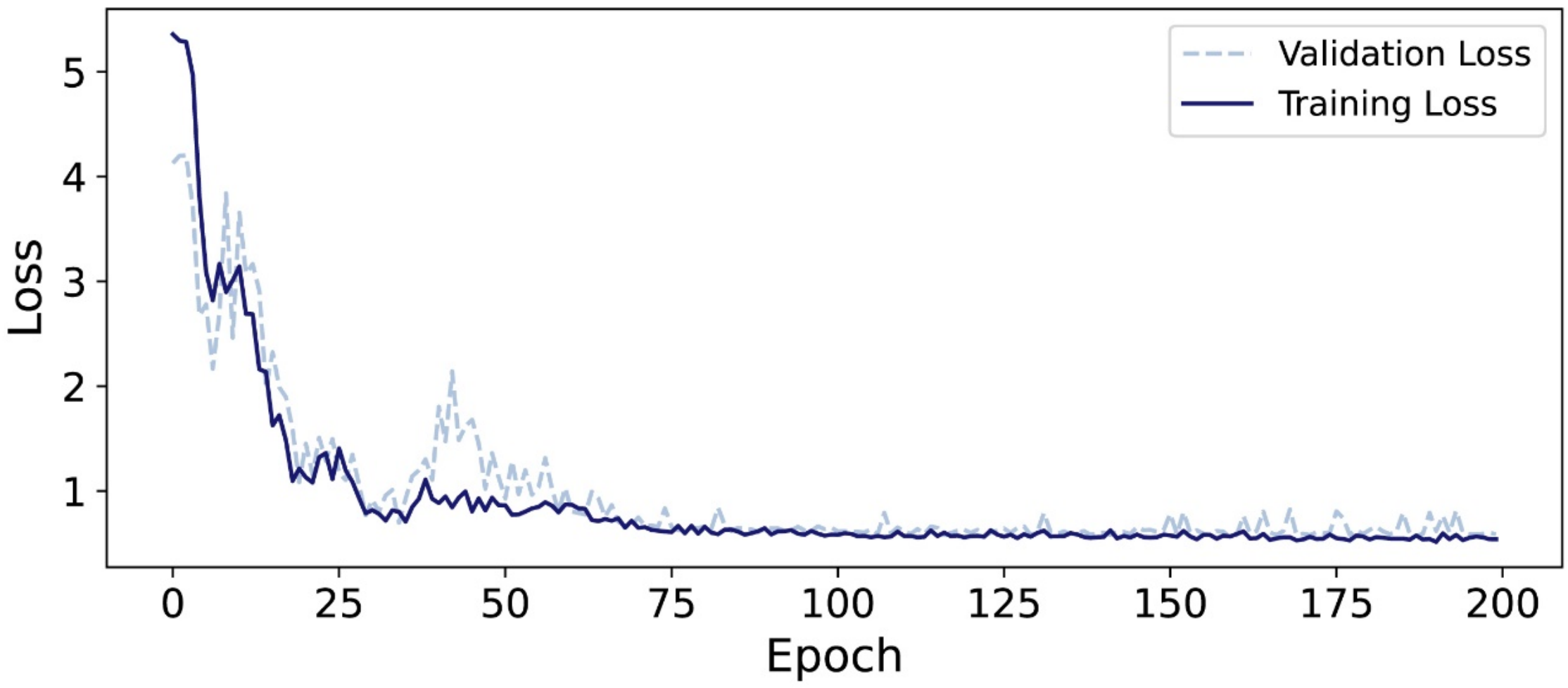

**Figure S7.** Training history of multi-class classification based on BoW model for **(a)** pore volume and **(b)** $CO_2$ Henry's constant datasets.

**Note S4: Influence of Circuit Depth in the Multi-Qubit Representations**

As discussed in the main manuscript, the construction of quantum circuits is differently handled based on the number of qubits representing each component. For the single-qubit representation per component (as in Figure 3b), the software allows multiple parameterized gates to be applied to each qubit. This is controlled via n_single_qubit_params argument, enabling users to specify number of parameters to represent each component using defined sequences of quantum gates such as Rx-Rz-Rx arrangement. This setting explicitly mitigates potential degeneracy issues (expressivity limitations) by enhancing each single-qubit's parameterization. In contrast, when a component is represented by multiple qubits (as in Figure 5a and Figure S8), lambeq automatically constructs a predefined entangled circuit architecture using Hadamard and controlled gates. In this case, lambeq systematically disables multiple parameterized single-qubit rotations, assigning only one trainable gate per qubit. This approach prevents over-parameterization and ensures well-defined entanglement patterns across the qubits.

To verify this behavior, we directly consulted Dr. Dimitri Kartsaklis, a co-developer of the lambeq package[2] and author of foundational QNLP literature[3], regarding this design choice (https://github.com/CQCL/lambeq/issues/131). He clarified that "The reason for this is that while any ansatz provides a way to combine two or more qubit into a meaningful circuit, it usually becomes degenerate in the single qubit case. This needs to be handled explicitly, which we do through the n_single_qubit_params argument." In other words, for single-qubit representations, explicit multi-parameter configurations (e.g. Rx-Rz-Rx) are necessary to avoid degeneracy, while for multi-qubit systems, degeneracy is naturally handled through circuit entanglement and fixed structure.

Since the lambeq adopted fixed circuit structure for multi-qubit representation, the number of parameters must be controlled by explicitly replicating the circuit depth as shown in Fig. S8b. To further explore whether increased parameterization in the multi-classification setting could improve performance, we conducted additional experiments by increasing the circuit depth from 1 layer (24 parameters, Figure S8a) to 3 layers (84 parameters, Figure S8b). The resulting training, validation, and test accuracies for both the pore volume and $CO_2$ Henry's constant dataset are shown in Figure S8. While the deeper circuit successfully increased the number of trainable parameters, the results did not yield improved accuracy. In fact, the 1-layer circuit consistently outperformed the 3-layer circuit across both datasets, with faster convergence (Figure S9a,c) and better performance (Figure S9b,d). This suggests that deeper circuits, while theoretically more expressive, may encounter optimization difficulties or overfitting when trained on limited-size datasets on NISQ-simulated environments.

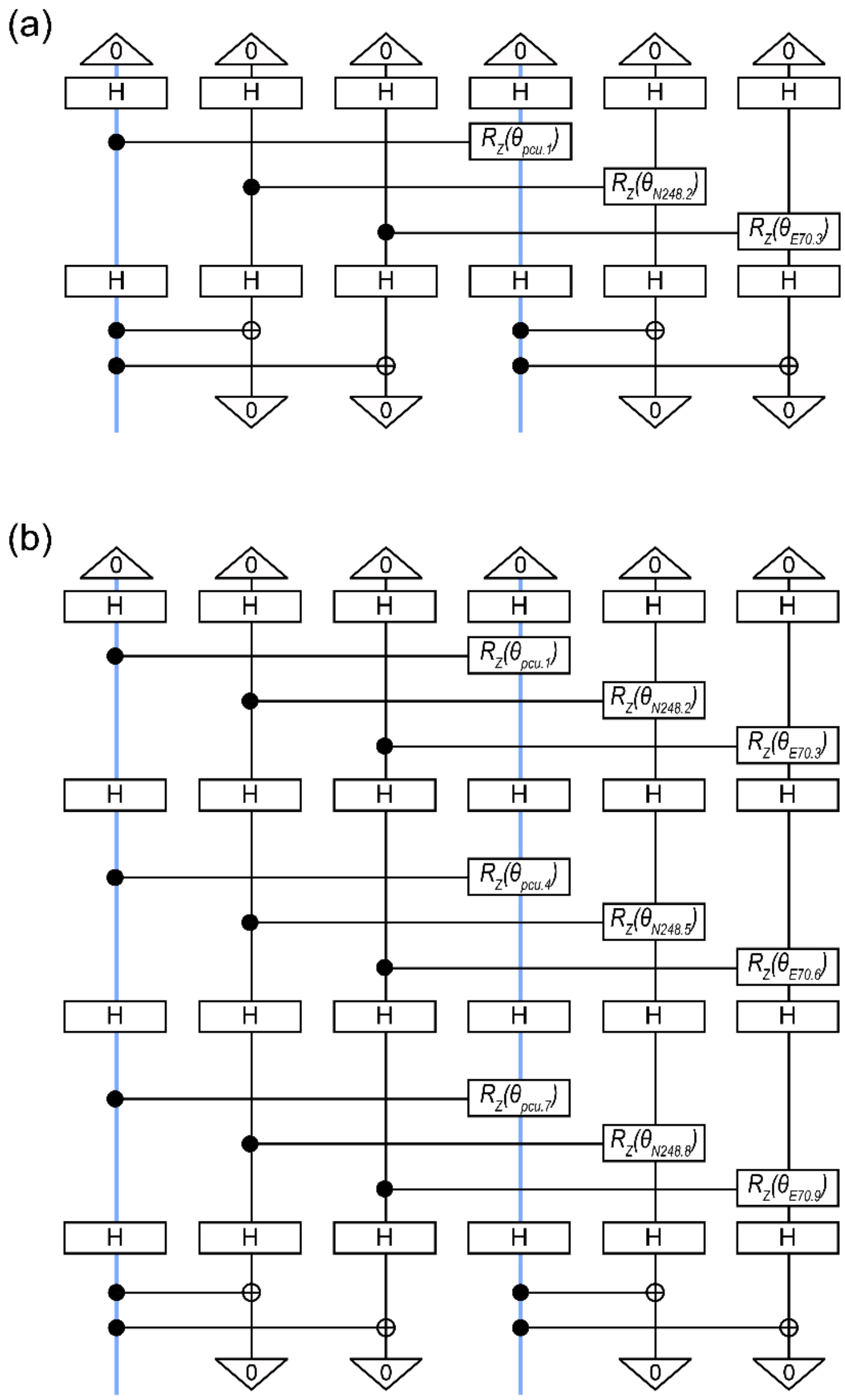

**Figure S8.** Quantum circuits for bog-of-words (BoW) QNLP model with varying circuit depths. **(a)** Circuit representation for 1-layer BoW model used in the original multi-classification task (as shown in Figure 5a), consisting of a single IQP layer with 24 trainable parameters (one controlled-Rz gate per qubit). **(b)** 3-layer circuit design with 84 trainable parameters, implemented to investigate the impact of circuit depth on model performance.

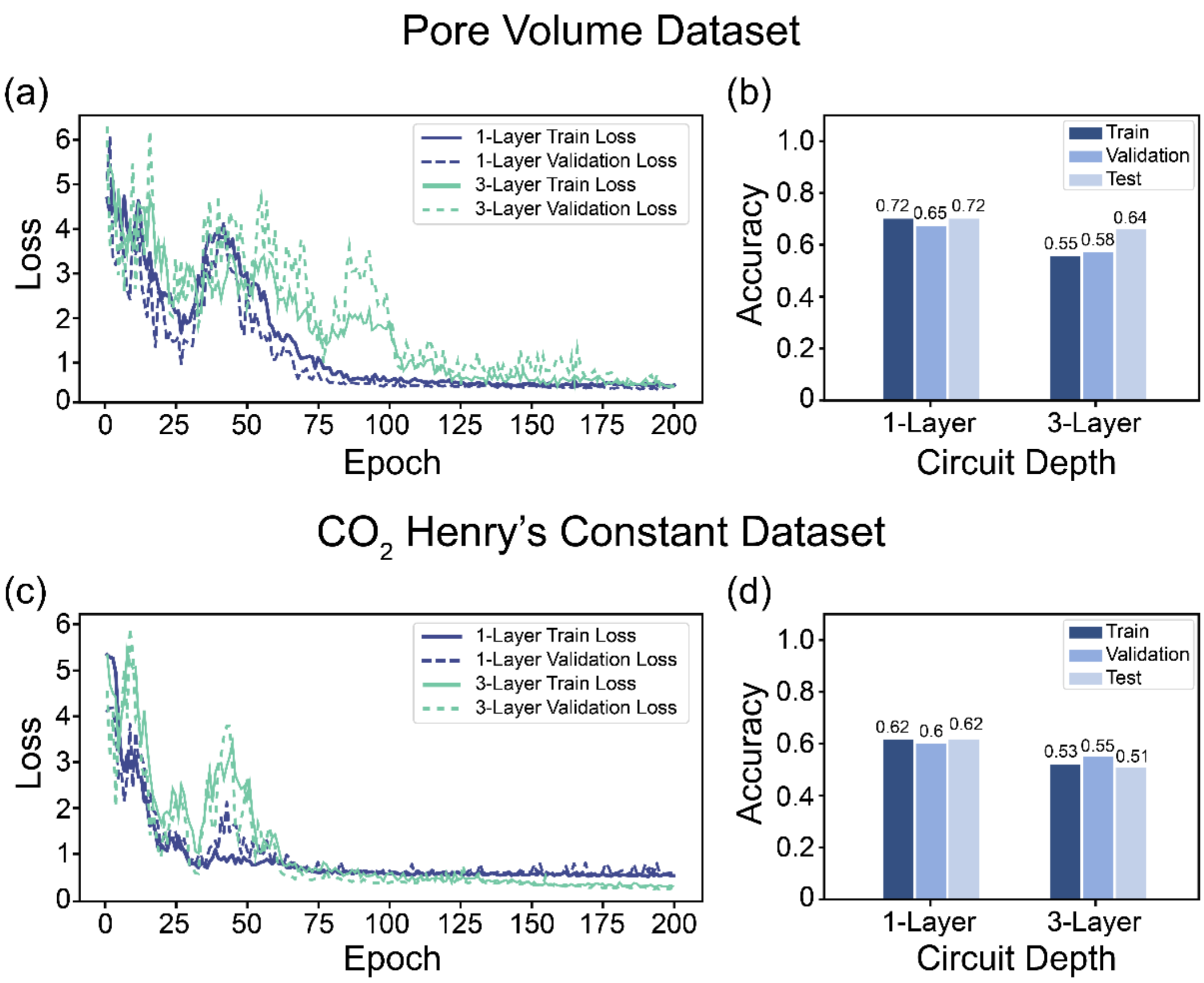

**Figure S9.** Comparison of training dynamics and classification accuracy between 1-layer and 3-layer BoW circuits. Evolution of training and validation losses for **(a)** the pore volume and **(c)** $CO_2$ Henry's constant datasets. Final classification accuracies for training, validation, and test sets using 1-layer and 3-layer circuits for **(b)** the pore volume and **(d)** $CO_2$ Henry's constant datasets.

**Table S3.** Comparison of evaluation metrics of QNLP models. Note that Multi, $Bi_{00}$, $Bi_{01}$, $Bi_{10}$, and $Bi_{11}$ represent BoW models trained based on the multi-classification dataset, 00-specialized dataset (binary classification dataset divided into classes 00 and 01+10+11), 01, 10, and 11-specialized datasets, respectively.

| **Property** | **Model** | **Accuracy** | | **Test** | | |
|---|---|---|---|---|---|---|
| | | **Training** | **Test** | **Precision** | **Recall** | **$F_1$** |
| **Pore Volume** | $Bi_{00}$ | 0.93 | 0.92 | 0.92 | 0.89 | 0.90 |
| | $Bi_{01}$ | 0.87 | 0.92 | 0.88 | 0.92 | 0.90 |
| | $Bi_{10}$ | 0.86 | 0.86 | 0.80 | 0.78 | 0.79 |
| | $Bi_{11}$ | 0.96 | 0.98 | 0.97 | 0.99 | 0.98 |
| | Multi | 0.50 | 0.50 | 0.39 | 0.50 | 0.40 |
| **$CO_2$ Henry's Constant** | $Bi_{00}$ | 0.87 | 0.86 | 0.80 | 0.78 | 0.79 |
| | $Bi_{01}$ | 0.83 | 0.72 | 0.66 | 0.63 | 0.64 |
| | $Bi_{10}$ | 0.80 | 0.78 | 0.69 | 0.66 | 0.67 |
| | $Bi_{11}$ | 0.86 | 0.84 | 0.82 | 0.74 | 0.77 |
| | Multi | 0.46 | 0.44 | 0.50 | 0.44 | 0.42 |

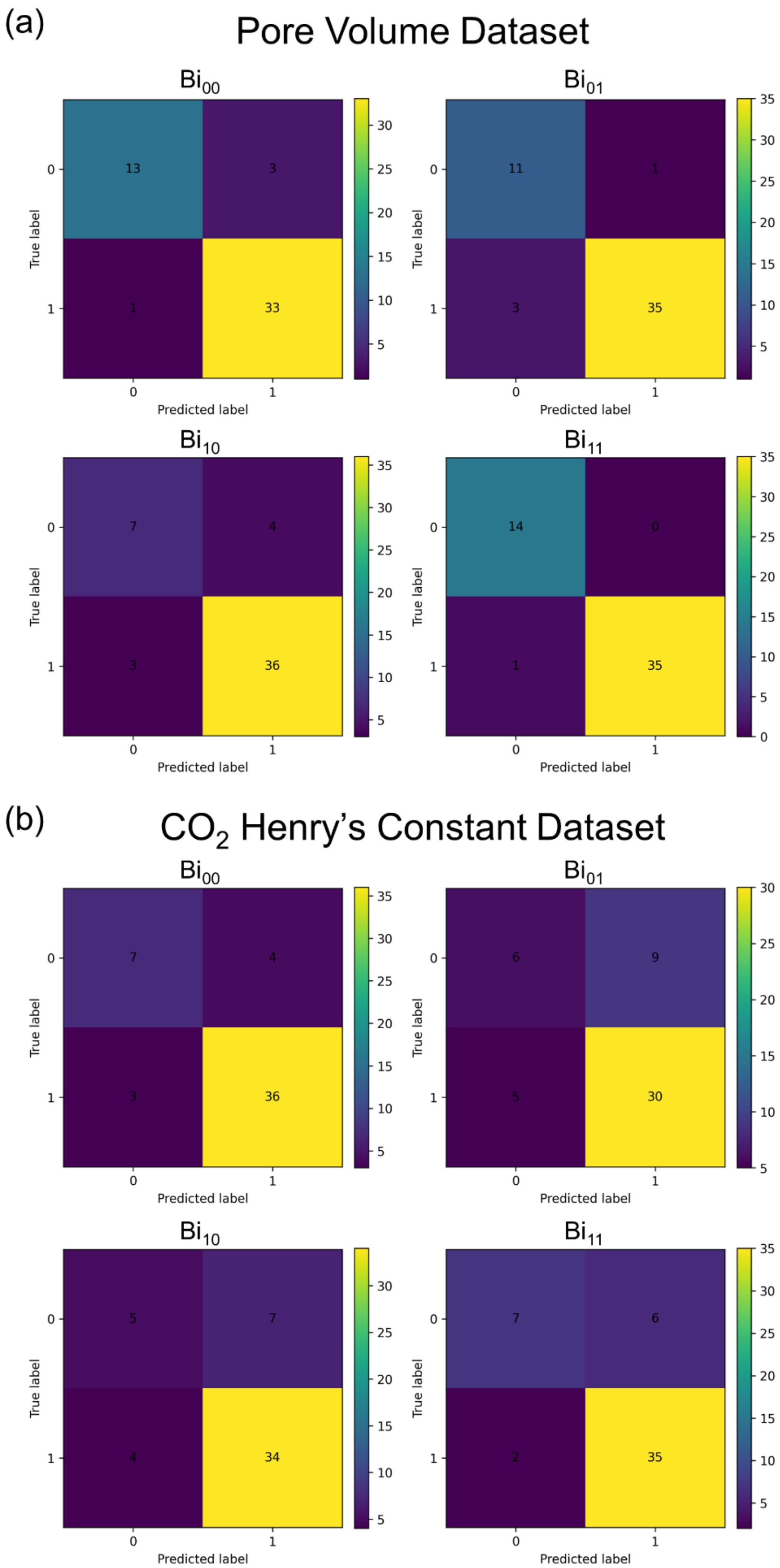

**Figure S10.** Confusion matrix of QNLP models on (a) pore volume and (b) $CO_2$ Henry's constant test datasets. $Bi_{00}$, $Bi_{01}$, $Bi_{10}$, and $Bi_{11}$ represent QNLP BoW models trained based on the 00-specialized dataset (binary classification dataset divided into classes 00 and 01+10+11), 01, 10, and 11-specialized datasets, respectively.

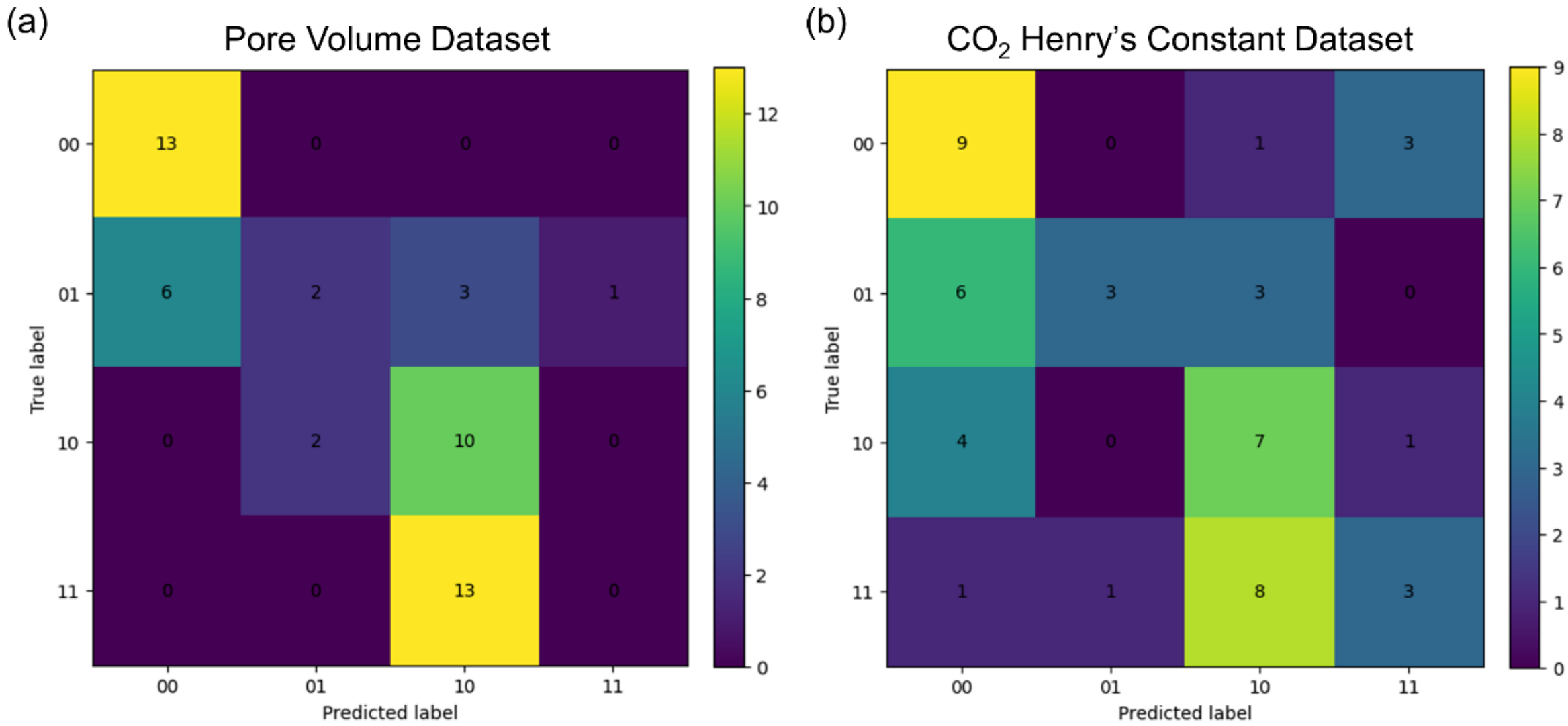

**Figure S11.** Confusion matrix of QNLP models for multi-class classification task on (a) pore volume and (b) $CO_2$ Henry's constant test datasets.

**Note S5: Optimization of Relative Probability Thresholds for Inverse Design Performance**

The comparative calculations were performed by varying threshold cutoffs of the relative probability for each target property, pore volume and $CO_2$ Henry's constants, to identify the best relative probability threshold that optimizes the inverse design performance. Tested threshold values were differently applied to account for the differences in dataset complexity. Thus, the threshold was ranged from 0.9 to 0.8 for search pace of the pore volume while 0.7 to 0.6 for that of $CO_2$ Henry's constants. In this setup, the system was forced to generate candidate MOFs exclusively from previously unseen combinations.

As summarized in Table S3, there was a significant difference in inverse design performance as the relative probability threshold changes. For the inverse design of MOFs targeting pore volume, a threshold of 0.85 yielded the best overall inverse design performance as the performance of moderately high class achieved 91% accuracy. Lowering the threshold to 0.8 degraded the accuracy for this class as 73%, while performance for other class remained nearly perfect. This decline aligns with the classification performance of the $Bi_{10}$ model, which had the lowest test accuracy of 86% among the binary classifiers as shown in Fig. 6b of the revised manuscript. This can be understood as the model is more likely to misclassify when the threshold drops below this classifier's performance, causing the design loop to fail more frequently or converge to incorrect solutions. For $CO_2$ Henry's constant, a threshold of 0.65 was found to offer the best overall performance, achieving a balanced accuracy across all classes while minimizing incorrect predictions and timeouts. Specifically, it maximized the accuracy for the moderately low class as 81% compared to only 46% at a threshold of 0.6. As the threshold increased, the accuracy for high and low classes improved steadily, while the number of timeouts for the moderately low and moderately high classes increased, reflecting the models' cautious behavior under stricter decision criteria. As discussed in the main manuscript, timeouts suggest the model withheld prediction due to insufficient confidence. Although the high number of timeouts at threshold 0.9 for the pore volume and 0.7 for $CO_2$ Henry's constant may preserve the possibility of eventual success, we concluded the optimal threshold for each target property as 0.85 and 0.65, considering the efficiency of the MOF generation process.

**Table S4.** Inverse design performance for generating MOFs with target property classes using QNLP classifiers. Each row summarizes the results of 100 independent generation tasks for a given property class, and relative probability threshold. The columns report the number of successful predictions (Correct), incorrect predictions (Incorrect), and instance where the model failed to produce a confident prediction within 100 iterations (Timeout). The Avg. guess column indicates the average number of attempts required to generate a successful MOF. The selected threshold values for this study were highlighted in gray color.

| Property | Threshold | Target Class | Correct | Incorrect | Timeout | Avg. guess |
|---|---|---|---|---|---|---|
| Pore volume | 0.9 | low | 100 | 0 | 0 | 13.71 |
| | | mod. low | 97 | 0 | 3 | 26.23 |
| | | mod. high | 98 | 1 | 1 | 27.15 |
| | | high | 87 | 0 | 13 | 40.38 |
| | 0.85 | low | 100 | 0 | 0 | 7.71 |
| | | mod. low | 100 | 0 | 0 | 10.73 |
| | | mod. high | 91 | 9 | 0 | 9.98 |
| | | high | 100 | 0 | 0 | 11.76 |
| | 0.8 | low | 100 | 0 | 0 | 4.84 |
| | | mod. low | 100 | 0 | 0 | 9.14 |
| | | mod. high | 73 | 27 | 0 | 7.24 |
| | | high | 100 | 0 | 0 | 8.36 |
| $CO_2$ Henry's constant | 0.7 | low | 92 | 8 | 0 | 7.44 |
| | | mod. low | 51 | 1 | 48 | 71.92 |
| | | mod. high | 95 | 0 | 5 | 33.66 |
| | | high | 98 | 2 | 0 | 9.27 |
| | 0.65 | low | 91 | 9 | 0 | 7.23 |
| | | mod. low | 81 | 16 | 3 | 28.44 |
| | | mod. high | 98 | 0 | 2 | 25.13 |
| | | high | 90 | 10 | 0 | 7.33 |
| | 0.6 | low | 90 | 10 | 0 | 6.37 |
| | | mod. low | 46 | 54 | 0 | 11.18 |
| | | mod. high | 95 | 5 | 0 | 18.02 |
| | | high | 83 | 17 | 0 | 5.15 |

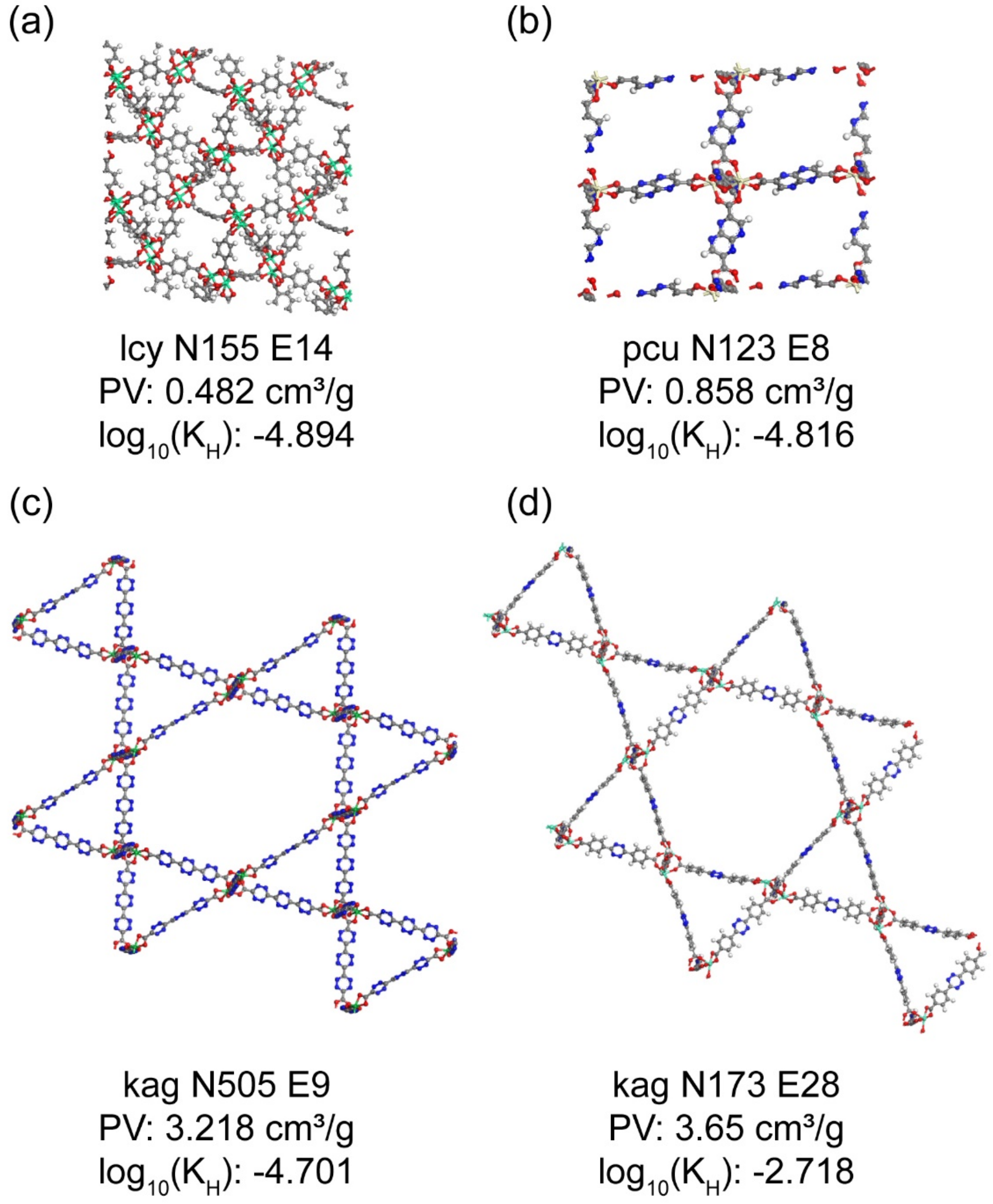

**Figure S12.** MOF structures that have **(a)** low, **(b)** moderately low, **(c)** moderately high, and **(d)** high pore volume and $CO_2$ Henry's constant (The unit of $CO_2$ Henry's constants is mol/kg/Pa).

**Table S5.** Results of varying the stability constant, A, based on pore volume and $CO_2$ Henry's constant datasets. The selected parameter setup for the model was highlighted in gray color.

| | A | metric | Model 00 | | Model 01 | | Model 10 | | Model 11 | |
|---|---|---|---|---|---|---|---|---|---|---|
| | | | train | validation | train | validation | train | validation | train | validation |
| Pore volume | 20 | loss | 0.234 | 0.149 | 0.367 | 0.396 | 0.354 | 0.34 | 0.194 | 0.213 |
| | | accuracy | 0.91 | 0.97 | 0.867 | 0.88 | 0.853 | 0.85 | 0.937 | 0.91 |
| | 2 | loss | 0.209 | 0.145 | 0.369 | 0.373 | 0.358 | 0.33 | 0.184 | 0.222 |
| | | accuracy | 0.907 | 0.97 | 0.842 | 0.86 | 0.835 | 0.835 | 0.907 | 0.92 |
| | 0.2 | loss | 0.205 | 0.133 | 0.372 | 0.391 | 0.337 | 0.349 | 0.188 | 0.226 |
| | | accuracy | 0.9 | 0.96 | 0.845 | 0.895 | 0.832 | 0.85 | 0.933 | 0.89 |
| $CO_2$ Henry's constant | 20 | loss | 0.399 | 0.446 | 0.419 | 0.403 | 0.463 | 0.442 | 0.385 | 0.33 |
| | | accuracy | 0.837 | 0.81 | 0.787 | 0.75 | 0.787 | 0.78 | 0.82 | 0.85 |
| | 2 | loss | 0.348 | 0.42 | 0.437 | 0.389 | 0.496 | 0.5 | 0.39 | 0.315 |
| | | accuracy | 0.823 | 0.84 | 0.818 | 0.8 | 0.808 | 0.78 | 0.823 | 0.9 |
| | 0.2 | loss | 0.382 | 0.398 | 0.43 | 0.43 | 0.458 | 0.4 | 0.387 | 0.304 |
| | | accuracy | 0.855 | 0.82 | 0.8 | 0.75 | 0.793 | 0.81 | 0.823 | 0.85 |